\documentclass[10pt,twocolumn,letterpaper]{article}

\usepackage[pagenumbers]{cvpr} %

\usepackage{graphicx}
\usepackage{amsmath}
\usepackage{amssymb}
\usepackage{booktabs}
\usepackage[accsupp]{axessibility}  %
\usepackage[pagebackref,breaklinks,colorlinks]{hyperref}
\usepackage{orcidlink}
\usepackage{soul}

\usepackage[dvipsnames]{xcolor}

\usepackage{amsmath,amsfonts,bm, bbm}

\def\eqref#1{equation~\ref{#1}}

\def\1{\bm{1}}

\def\vc{{\bm{c}}}

\def\vp{{\bm{p}}}

\def\vs{{\bm{s}}}

\def\mC{{\bm{C}}}

\def\mP{{\bm{P}}}

\def\mS{{\bm{S}}}

\DeclareMathAlphabet{\mathsfit}{\encodingdefault}{\sfdefault}{m}{sl}
\SetMathAlphabet{\mathsfit}{bold}{\encodingdefault}{\sfdefault}{bx}{n}

\usepackage{gensymb}
\usepackage[dvipsnames]{xcolor}
\usepackage{multirow}
\usepackage{array, makecell}
\usepackage{amssymb}
\usepackage{pifont}
\newcommand{\xmark}{\ding{55}}%
\newcommand{\cmark}{\ding{51}}%
\usepackage{wrapfig}
\usepackage{multirow,bigstrut}
\usepackage{tabu}
\usepackage{algorithm}
\usepackage{algorithmic}

\usepackage[dvipsnames]{xcolor}

\definecolor{pose}{RGB}{212, 170, 0}
\definecolor{nearest}{RGB}{0, 255, 0}
\definecolor{cp}{RGB}{0,244,255}

\usepackage[capitalize]{cleveref}
\crefname{section}{Sec.}{Secs.}
\Crefname{section}{Section}{Sections}
\Crefname{table}{Table}{Tables}
\crefname{table}{Tab.}{Tabs.}

\def\wacvPaperID{173} %
\def\confName{WACV}
\def\confYear{2025}

\begin{document}

\title{GaitContour: Efficient Gait Recognition based on a Contour-Pose Representation}

\author{\parbox{16cm}{
\centering{Yuxiang Guo, Anshul Shah, Jiang Liu, Ayush Gupta, \\
Rama Chellappa,  Cheng Peng}\\
{Johns Hopkins University, Baltimore, MD, USA} \\
{\tt\small \{yguo87, ashah95, jiangliu, agupt120, rchella4, cpeng26 \}@jhu.edu}\\
}
}

\maketitle

\begin{abstract}
Gait recognition holds the promise to robustly identify subjects based on walking patterns instead of appearance information. In recent years, this field has been dominated by learning methods based on two input formats: silhouette images and sparse keypoints. Compared to image-based approaches, keypoint-based methods can achieve significantly higher efficiency due to their sparsity. However, sparsity also results in information loss, thereby reducing performance. In this work, we propose a novel, keypoint-based Contour-Pose representation, which compactly encodes both body shape and parts information. We further propose a local-to-global architecture, called GaitContour, to leverage this novel representation and efficiently compute subject embedding in two stages. The first stage consists of a local transformer that extracts features from five different body regions. The second stage then aggregates the regional features to estimate a global human gait representation. Such a design significantly reduces the complexity of the attention operation and improves both efficiency and performance. Through large scale experiments, GaitContour is shown to perform significantly better than previous keypoint-based methods. Furthermore, the Contour-Pose representation also achieves new SoTA performances on fusion-based gait recognition methods.
\vspace{-5pt}
\end{abstract}

\section{Introduction}
\label{sec:intro}

\begin{figure}%
  \vspace{-3mm}
    
    \scriptsize
    \setlength{\tabcolsep}{2pt}
    \centering
    \begin{tabular}{c| c| c| c}
    
        \multicolumn{1}{c}{} & 
        \multicolumn{1}{c}{
        \begin{subfigure}[c]{.24\linewidth}
        \centering
            \includegraphics[width=\textwidth,height=\textwidth]{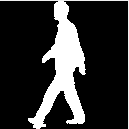}
            \caption*{Silhouette}
            \label{sil}
        \end{subfigure} 
        } &
        \multicolumn{1}{c}{
        \begin{subfigure}[c]{.24\linewidth}
        \centering
            \includegraphics[width=.5\textwidth,height=\textwidth]{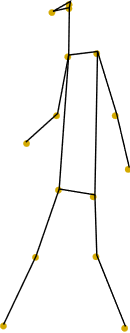}
            \caption*{Pose}
            \label{ske}
        \end{subfigure}
        } &
        \multicolumn{1}{c}{
        \begin{subfigure}[c]{.24\linewidth}
        \centering
            \includegraphics[width=.5\textwidth,height=\textwidth]{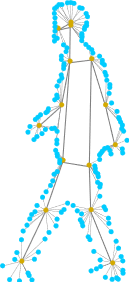}
            \caption*{{CP}}
            \label{conpo}
        \end{subfigure}
        }\\
    \toprule
        Computation & High & Low & Low \\
        Format & Image & Keypoint & Keypoint\\
        Body Shape & \cmark & \xmark &\cmark\\
        Size & $64\times 64$ & $17\times 2$ & $165 \times 2$ \\
        Model & GaitBase~\cite{fan2023opengait} & GPGait~\cite{fu2023gpgait} & GaitContour \\
        Parameters & 7.30M & 7.93M & \textbf{0.66M \textcolor{red}{$\downarrow$}\textcolor{red}{\tiny{$\times 10+$}}}\\
    \bottomrule
    \end{tabular}
    
    \centering
    \includegraphics[width=\linewidth]{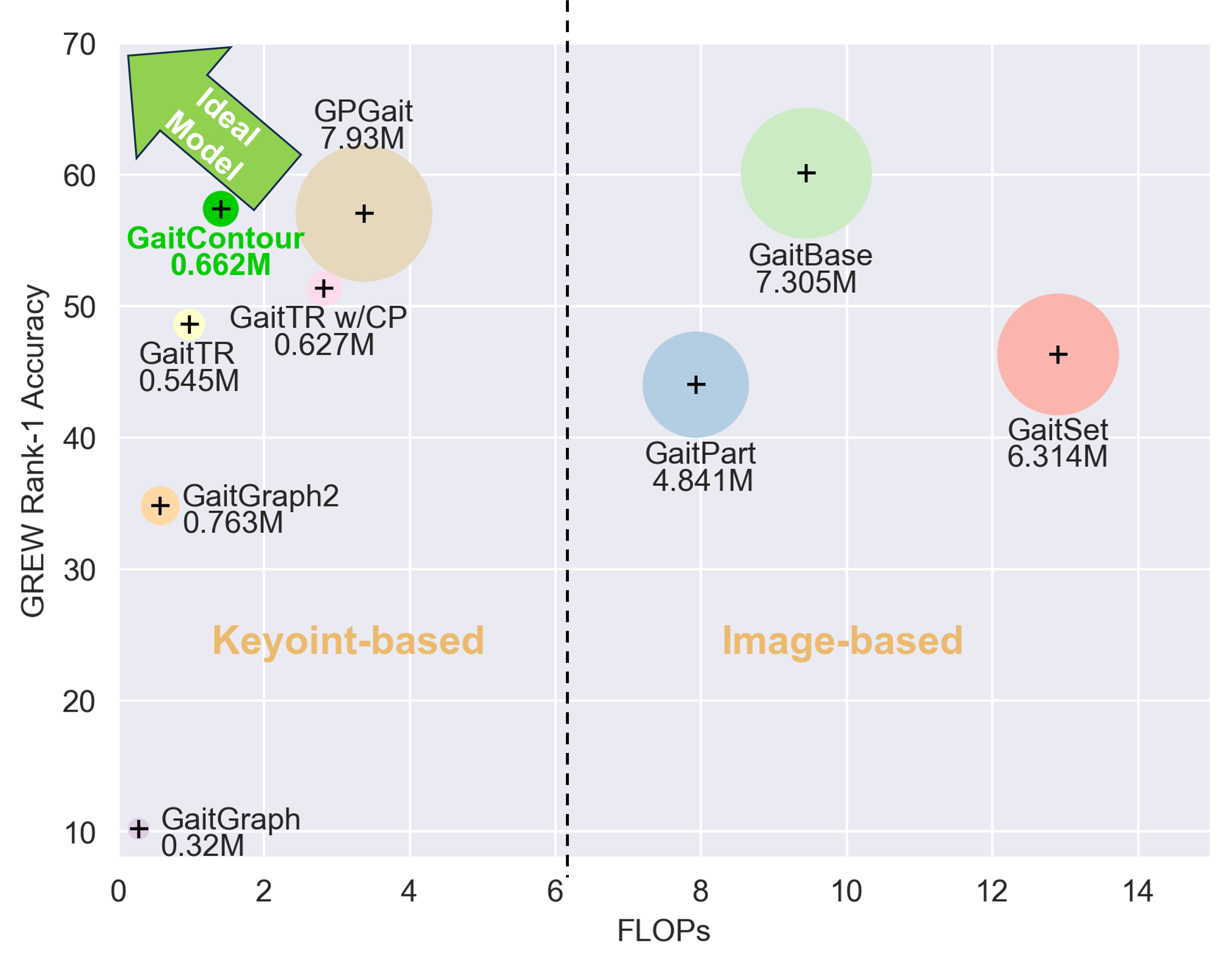}
    \caption{Comparison of our proposed Contour-Pose and other gait representations. The size of bubbles denotes the number of parameters. GaitTR w/CP represents extracting Contour-Pose(CP) feature through GaitTR. GaitContour achieves a good balance between efficiency and accuracy.}
    \label{fig:comparison}
  \vspace{-6mm}
\end{figure}

Unconstrained biometric identification, especially in outdoor and long-range situations, has been a longstanding challenge~\cite{zhu2021gait,zheng2022gait,sepas2022deep,shen2022comprehensive}. While RGB-based face and body recognition systems focus on learning \emph{spatially} discriminative features, real-world effects like challenging viewpoints, low face resolution, changing appearances (\eg, clothes and glasses), \textit{etc.}, can significantly affect model performance~\cite{li2018resound,li2019repair,weinzaepfel2021mimetics}. Gait analysis employs an alternative modality for human recognition by learning discriminative features extracted from human walking patterns. It can be more robust in challenging, unconstrained situations, where color-space information is unreliable due to turbulence, and has been deployed in many applications including human authentication~\cite{benedek2016lidar}, health~\cite{del2019gait}, crime analysis~\cite{hadid2012can}, \textit{etc.} 

Research on gait analysis has a long history~\cite{9714177,shen2022comprehensive}. The more recent developments in this field are based on deep learning methods, where the format of inputs to the neural network can be roughly categorized in two ways: silhouette images and keypoints. To extract discriminative features from these two types of data, models of different complexities are applied. Previous works mainly focus on the performance differences from a modality perspective, such as silhouette, keypoints, or multi-modal. This study analyzes the \textbf{models' efficiency and effectiveness tradeoff} by using different representations, which is crucial for assessing the implementation cost in real-life applications.

Gait image sequences, which capture human motion as a series of 2D binary silhouettes or skeleton maps, are typically processed by large CNN models. Relying on dense representations, these models require significant computations to generate effective features, which often lead to a latency of 80-100ms. Such cost makes image-based gait recognition less favorable compared to face and body recognition, which takes around 10-20ms, despite its advantages in privacy preservation and performance.

Pose keypoints are predefined semantic points extracted from human images and magnitudes smaller in input size than images. Such smaller dimensionality offers several benefits, including smaller models, faster processing, and smaller template sizes (the identification vectors generated by the model), However, the reduced accuracy of 
keypoint-based models is a major bottleneck. 
If keypoint-based models can achieve competitive performance, they can enable broader downstream applications, e.g., gait recognition in low-power, efficient systems with real-time demands.

Considering the advantages and disadvantages of the two gait representations, we pose this question: can we use slightly denser keypoints to represent human movement and improve keypoint-based models' performance? Based on the success of the image-based models, we argue that body shape also plays a significant role in gait recognition. As such, it can be beneficial to extract more keypoints around the contour of a human to complement semantic human poses. 
To this end, we propose a novel gait representation called Contour-Pose, as illustrated in~\cref{fig:comparison}. We 
compactly represent the human body shape using a series of contour points around the silhouette, \eg, approximated by Teh-Chin algorithm~\cite{teh1989detection}. However, naively using contour points cannot achieve good performance due to the inconsistent correspondence and ordering between frames.
To address this issue, we draw inspiration from the blend skinning process in building the Skinned Multi-Person Linear Model (SMPL)~\cite{loper2023smpl}, where the final skin vertex locations are highly correlated to the joint centers. We use pose keypoints as anchors to select relevant contour points. Specifically, for every pose keypoint, we select a few close contour points to form a connected graph in a clockwise fashion, simulating the SMPL skinning process in a 2D style.
Together, this contour-pose representation compactly represents the semantic regions of the human body and its shape. Compared to the typical silhouette, our Contour-Pose representation is an order of magnitude smaller in dimension.

When used in place of conventional pose keypoints, Contour-Pose can already improve the performances of prior keypoint-based gait recognition models~\cite{zhang2023spatial, teepe2021gaitgraph, teepe2022towards}. 
However, these models are designed for sparse pose keypoints, and incur higher computational costs due to the larger amount of point inputs in Contour-Pose and the quadratic complexity of Transformers, which are commonly used in motion analysis~\cite{zhang2023spatial, plizzari2021spatial,bello2019attention}. Our analysis reveals that a significant portion of the computation in a Transformer is dedicated to relationships that are not usually pertinent, \eg, there is little correlation between the contour points surrounding the head and legs. 

In light of this consideration, we propose GaitContour, a Transformer-based method that is designed in a local-to-global fashion to maximize performance and efficiency. GaitContour operates in two stages: a Local Contour-Pose Transformer (Local-CPT), and a Global Pose-Feature Transformer (Global-PFT). As contour points are defined with respect to keypoints, we propose a Local-CPT to extract local features of the specific regions using shared weights. This design reduces the number of model parameters and allows for sharing the general low-level features. Local-CPT's outputs are aggregated to form global keypoint features. The Global-PFT focuses on this sparse set of global keypoint features to generate human IDs.

As shown in ~\cref{fig:comparison}, GaitContour achieves a significantly better efficiency-performance trade-off based on richer information and tailored architecture design, compared to previous keypoint-based models~\cite{fu2023gpgait}. Furthermore, we can leverage Contour-Pose in place of keypoints for image-keypoint-fusion modeling, similar to the setup of SkeletonGait++~\cite{fan2023skeletongait}, and achieve new State-of-The-Art results. This further demonstrates the effectiveness of our proposed representation. In summary, our contributions are as follows:

\begin{enumerate}
\item We propose a novel gait representation, called Contour-Pose, which augments pose keypoints with contour points extracted from silhouettes; this representation contains rich information, is compact in size, and can improve current keypoint-based gait recognition methods.
\item We propose a novel gait recognition method, called GaitContour, which leverages Contour-Pose and a Transformer-based design; GaitContour processes Contour-Pose in a local-to-global fashion, which maximizes efficiency.
\item We evaluate our novel gait representation and recognition method over several large-scale datasets, and find significant performance and efficiency improvements compared to previous SOTA methods both in keypoint-only and fusion scenarios.
\end{enumerate}

\section{Related Works}
\label{sec:related_work}

\subsection{Gait Representation}
\label{subsec:gait_representation}
In past decades, researchers have used a variety of representations to capture human gait motion, including RGB images~\cite{zhang2020learning, li2020end}, binary masks/silhouettes~\cite{chao2019gaitset,fan2020gaitpart,lin2021gait}, optical flow images~\cite{Dosovitskiy_2015_ICCV,ilg2017flownet,hui2018liteflownet}, 2D skeleton/pose keypoints~\cite{fu2023gpgait,teepe2021gaitgraph,teepe2022towards,zhang2023spatial}, and gait-oriented templates, like Gait Energy Image(GEI)~\cite{han2005individual}, Gait History Image~\cite{bobick2001recognition}, \textit{etc}. In recent years, novel gait representations have also emerged, in the form of Li-DAR pointcloud~\cite{shen2023lidargait}, 3D mesh~\cite{li2020end,zheng2022gait}, and event stream cameras~\cite{wang2019ev}; but these representations are difficult to compute, and consequently, datasets are limited in scale. 
The current datasets and methods are mainly based on \textit{images} and \textit{keypoints}, which are the focus of this work.

\subsection{Keypoint-based Gait Recognition}
\label{subsec:model_based}
Keypoint-based methods are typically designed to predict identities based on 2D pose keypoints across many frames. These methods utilize semantic positions, \eg,
knees and wrists, along with their spatial and temporal connections as inputs. By distilling images into semantically meaningful points, keypoint-based methods can be more robust to noisy factors such as different clothing and self-occlusion. GaitGraph~\cite{teepe2021gaitgraph} and its successor GaitGraph2~\cite{teepe2022towards} treat pose keypoints as a graph and employ a Graph Convolutional Network (GCN) to extract features. GaitTR~\cite{zhang2023spatial} and GaitMixer~\cite{pinyoanuntapong2023gaitmixer} capture the global temporal and spatial relationship through a transformer-like~\cite{vaswani2017attention} architecture. GPGait~\cite{fu2023gpgait} is based on a Part-Aware Graph Convolutional Network (PAGCN) to explore the keypoint representation under cross-domain settings.~\cite{guo2023physics} applies physics-augmented autoencoder to distill physics knowledge into gait recognition. Even though keypoints perform well in other motion-related tasks such as action recognition~\cite{yan2018spatial}, they have lower performance in gait recognition compared to image-based methods; but they are generally much faster and more efficient because of sparser inputs.

\subsection{Image and Fusion-based Gait Recognition}
\label{subsec:appearance_based}
The human silhouette is a mainstream representation used in current image-based recognition methods, typically based on Convolutional Neural Networks (CNNs). Specifically, 
GaitGL~\cite{lin2021gait} improves the quality of embeddings further by aggregating both local and global descriptors to capture local details and contextual relations. Recent works have explored different backbones, \eg, GaitBase~\cite{fan2023opengait}, DeepGaitV2~\cite{fan2023exploring}, to achieve higher performance across diverse datasets, especially in outdoor unconstrained scenarios, \eg, GREW~\cite{zheng2022gait}. Recently, SkeletonGait~\cite{fan2023skeletongait} transferred the keypoints into a skeleton map, so that a large feature extraction model can be employed, achieving higher performance. Comparatively, image-based methods take longer to process, e.g. 102ms/sequence for DeepGaitV2~\cite{fan2023exploring}, as they perform convolution operations at every pixel of the sequence, and consist of many layers. 

Given the unique characteristics inherent to each modality, several works focus on fusing the inputs. Castro et al.~\cite{castro2020multimodal} fuse depth map, gray image and optical flow through CNNs to perform gait recognition. DME~\cite{guo2023multi} incorporates RGB images and silhouettes, enriching the feature space expression. SMPLGait~\cite{zheng2022gait} integrates 2D and 3D modalities, i.e. silhouette and human mesh, in feature spaces, which have been shown to improve performance. Bifusion~\cite{peng2023learning} and MMGaitFormer~\cite{cui2023multi} fuse silhouette and pose information, employing concatenation and cross attention respectively. These methods primarily concentrate on fusing the modalities in feature spaces, as the extraction of different modalities involves multiple isolated backbones.

\section{Method}
\label{sec:method}

\subsection{Contour-Pose}
\label{subsection: CP}

\begin{figure*}[htb]
    \vspace{-9mm}
    \setlength{\abovecaptionskip}{3pt}
    \setlength{\tabcolsep}{2pt}
    \centering
    \includegraphics[width=\linewidth]{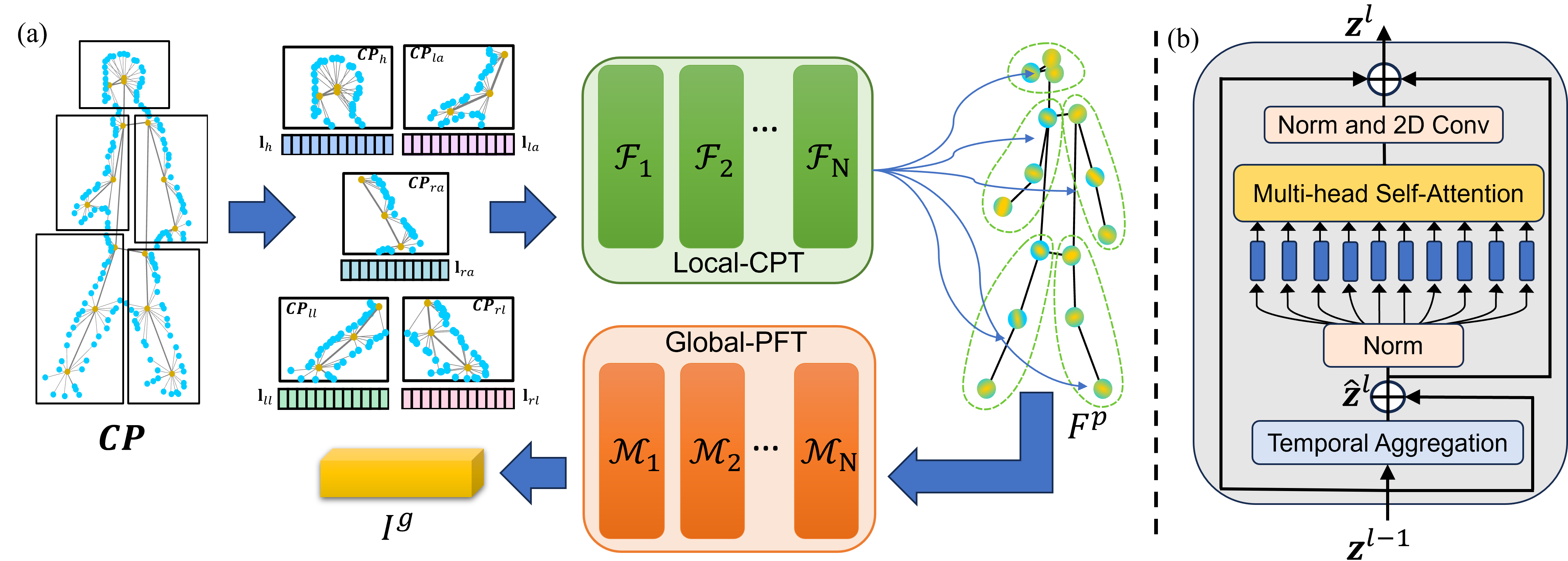}
    
    \caption{(a) \textbf{An Overview of GaitContour}. The Contour-Pose is partitioned into five regions, i.e. head, left arm, right arm, left leg, and right leg. Local-CPT extracts features from each region separately. GaitContour combines these local features into an identity embedding through a Global Pose-Feature Transformer. This local-to-global design enhances both efficiency and effectiveness for GaitContour. (b)\textbf{ The structure of the Temporal Transformer Layer}. It extracts the spatiotemporal correlation between each point, serving as a basic block for Local-CPT and Global-PFT.}
   \label{fig:network}
\vspace{-6mm}
\end{figure*}

Inspired by previous works that attempt to combine the body shape and pose in the feature space~\cite{peng2023learning,cui2023multi}, we look at a more principled approach to extract body shape features without any neural networks to design keypoint-based method with more representation fidelity. Ideally, this novel representation should have the following properties:

\begin{itemize}
    \item \textit{Information Preservation}: The representation should have minimal information loss. i.e., the original signals can be restored from the representation.
    \item \textit{Compactness}: The representation should be concise, such that its downstream processing is efficient.
    \item \textit{Temporal Consistency}: The representation should have consistent ordering across frames.
\end{itemize}

\begin{figure}
    \centering
    \includegraphics[width=1\linewidth]{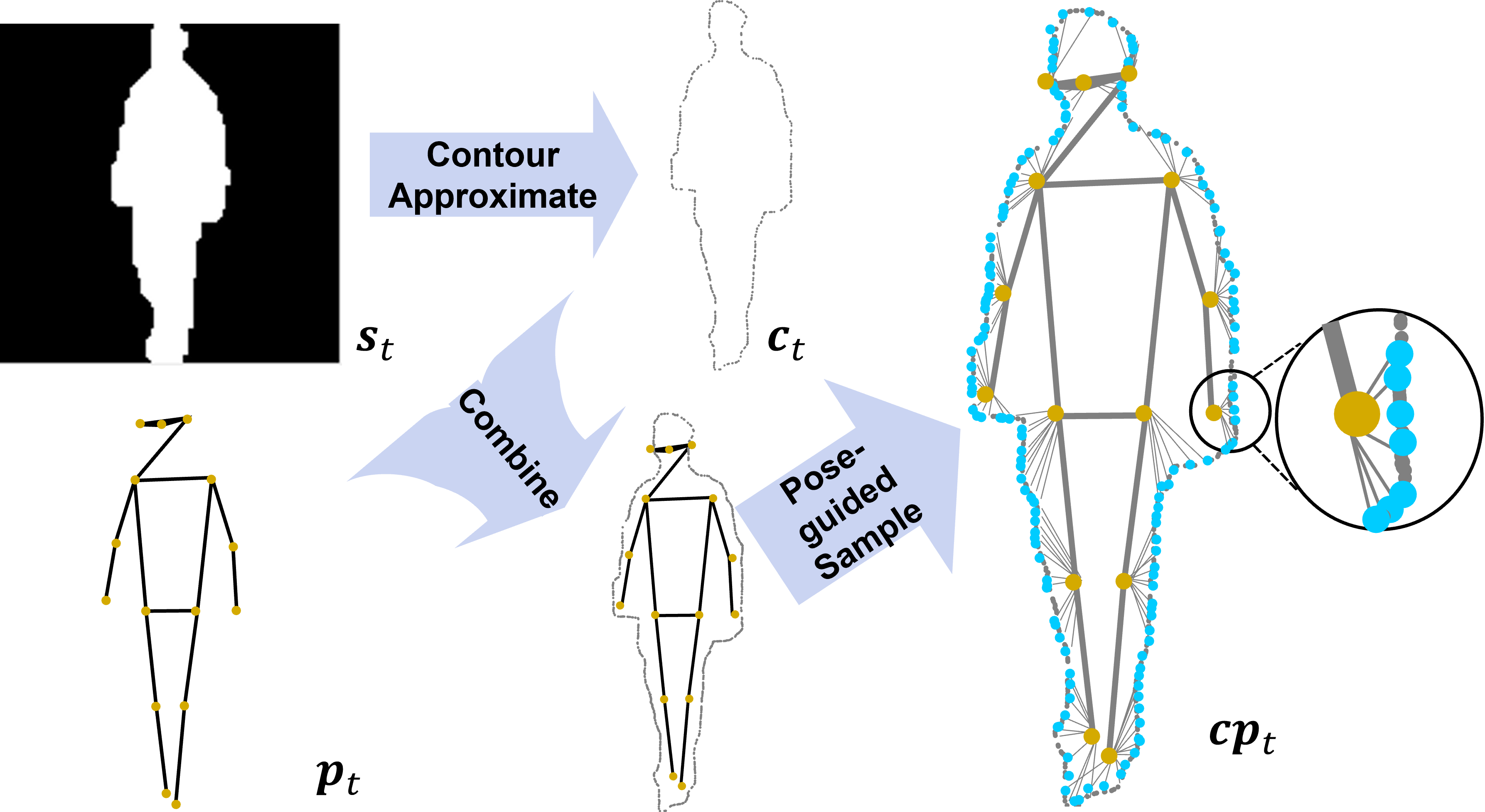}
    \caption{The construction of Contour-Pose. The pose is combined with the contour points sampled from the silhouette edge. In particular, contour points are sampled based on their distances from neighborhood poses. As shown in the zoomed area, \textcolor{cp}{Contour-Pose} is the $n$ nearby contour points from each \textcolor{pose}{pose} with connections. }%
    \label{fig:contourpose}
    \vspace{-5mm}
\end{figure}

Pose keypoints are very compact, but do not contain enough information to reconstruct the silhouettes; silhouettes contain more information, but are not as compact. To this end, we propose \textbf{Contour-Pose}, which compresses a silhouette to a series of contour points to augment pose keypoints. The contour points at sufficient density preserve most information in a mask, while being compact. The process to produce Contour-Pose is demonstrated in~\cref{fig:contourpose}, and is defined formally next. 

Suppose we have a subject's silhouette $\mS$ and pose $\mP$ across $T$ frames: 
\begin{align}
    \mS = [\vs_t \in \mathbb{R}^{H \times W}]_{t=1}^T, \mP = [\vp_t \in \mathbb{R}^{V \times 2}]_{t=1}^T ,
\end{align}
where $H, W$ and $V$ represent frame height, width, and the number of keypoints. For each frame, pose keypoints consist of $V$ nodes $[(x_1, y_1),...(x_V,y_V)]$ in 2D, and $E$ edges between them, which are commonly defined~\cite{teepe2021gaitgraph,zhang2023spatial}.

We hypothesize that a significant portion of the useful information in $\vs_t$ lies on edges, as similar ideas have been examined before~\cite{liang2022gaitedge, baumberg1994learning,wang2003silhouette}. Instead of using an edge image as previous works~\cite{liang2022gaitedge}, we sample the points on it, leading to a more compact representation to describe the body shape. To this end, we employ popular \textbf{contour approximation} methods~\cite{teh1989detection,opencv_library} to 1). compute the silhouette edge points, and 2). approximate the edge with a lower number of contour points $\vc_t$. In practice, 300+ points are extracted on the contour of $\vs_t$. %

Although the approximated contour compactly contains the silhouette information, in practice, we find that contour points alone do not yield good performance with current keypoint-based gait recognition methods~\cite{teepe2021gaitgraph,zhang2023spatial}. This is likely due to the lack of temporal consistency in contour point approximation, where the points in neighboring frames do not have the same semantic meaning and may arbitrarily shift based on the approximation algorithm. This consistency helps to build a stable graph to describe the walking pattern.

To establish consistent connections among frames, we draw inspiration from the blending skin process in SMPL~\cite{loper2023smpl}, where the skin vertices are controlled by the joint positions.  We treat the contour and pose points as analogous to skin and joints in 2D, leveraging pose keypoints as a \textbf{semantic guide} to refine contour point selection. For each pose point in $\vp_t$, we select $n$ nearby contour points to form a connected graph.
To ensure consistency, we arrange the contour points associated with each pose keypoint in clockwise order. The merged contour and pose keypoints, named Contour-Pose, are defined as follows:
\begin{align}
    \mC\mP = [\vc\vp_t \in \mathbb{R}^{(V\times n+V) \times 2}]_{t=1}^T.
\end{align}

As demonstrated in~\cref{fig:contourpose}, Contour-Pose combines the semantic information expressed in poses and the body shape information expressed by silhouettes. By directly applying Contour-Pose on current keypoint-based methods, \eg, GaitTR~\cite{zhang2023spatial}, we can already observe significant improvements in gait recognition performances without any architectural modification, as shown in~\Cref{tab:input} (d-e). For more details on the construction of Contour-Pose, please refer to the supplemental material.

\subsection{GaitContour}
\label{subsection: GaitContour}

Contour-Pose already improves the baseline gait recognition performance. We note that Transformer-based methods like GaitTR~\cite{zhang2023spatial} compute attention in $\mathcal{O}(n^2)$ complexity with respect to the input points. This leads to significantly more operations, as Contour-Pose has $V\times n$ additional points on top of pose keypoints. We propose \textbf{GaitContour}, an efficient transformer-based gait recognition model developed for Contour-Pose. As shown in~\cref{fig:network} (a), GaitContour first computes local features in five defined regions with a Local Contour-Pose Transformer (Local-CPT); it then aggregates local features and computes a global representation with a Global Pose-Feature Transformer (Global-PFT). Such a local-to-global design allows each transformer to focus on relevant points and features, thereby significantly improving efficiency and performance. Both Local-CPT and Global-PFT are built using the Temporal Transformer Layer.

\subsubsection{Temporal Transformer Layer}

To compute the spatiotemporal correlation between each point, we utilize a Temporal Transformer Layer (TTL) as shown in~\cref{fig:network} (b).  It captures the points' movement with time through Temporal Aggregation (TA) and uses a self-attention mechanism to extract the relations between each point. Each layer's operations can be described and formulated as follows:
\begin{equation}
\begin{aligned}
    &\hat{\textbf{z}}^l= {\textbf{z}}^{l-1} + \text{TA}({\textbf{z}}^{l-1}), \\ 
    & {\textbf{z}}^{l} = \text{BN}(\text{Conv}({\text{MHA}(\text{BN}(\hat{\textbf{z}}}^l))))+ \hat{\textbf{z}}^l+ {\textbf{z}}^{l-1},\\
\end{aligned}
\end{equation}
where $\hat{\textbf{z}}^l$, ${\textbf{z}}^{l}\in \mathbb{R}^{T\times J\times C}$ are the output features of TA and layer $l$, respectively; $T, J, C$ denotes the frame number, the number of points/tokens and the channel dimension. We follow~\cite{zhang2023spatial} to use convolutions along the time axis to reason about temporal information in the TA module. 
This temporal transformer structure is commonly used~\cite{zhang2023spatial, plizzari2021spatial,bello2019attention}, and is shown to be effective for gait analysis.

\subsubsection{Local Contour-Pose Transformer}

Contour points in $\mC\mP$ are defined in relationship to pose keypoints and represent more detailed body-shape information; To leverage this relationship, we propose to first compute local features based on contour points and keypoints. It stands to reason that local details, \eg, contour shapes in the left foot, bear a minor correlation to contour points surrounding the head. Guided by this hypothesis, we define five body regions based on keypoints - head, left arm, right arm, left leg, and right leg, defined as:
\begin{align}
    \mC\mP_{r} \in \mathbb{R}^{T\times (3+3n) \times 2},  \forall r \in \{h, la, ra, ll, rl\},
\end{align}
each region contains 3 keypoints and each keypoint is associated with $n$ contour points, a total of $(3+3n)$ points.

A Local Contour-Pose Transformer (Local-CPT) is a transformer architecture built upon TTL. Denoting every TTL block within Local-CPT to be $\mathcal{F}_i$, the forward process of Local-CPT can be described as follows:
\begin{gather}
    F_{r} = \mathcal{F}_N \circ \mathcal{F}_{N-1} \circ ... \mathcal{F}_{2}(\mathcal{F}_1(\gamma(\mC\mP_{r}))\oplus \mathbf{l}_{r})),\\
    F^{p}_{r} = AvgPool(F_{r}, n+1),
\end{gather}
\noindent where $\circ$ represents operate in series, $\mathbf{l}_{r}\in \mathbb{R}^{1\times C_{1}}$ is a learnable regional embedding indicating the current Contour-Pose region, $F_{r} \in \mathbb{R}^{T\times (3+3n) \times C_{N}}$ is the output feature vector for every region, $C_{i}$ is the channel size of $\mathcal{F}_i$ and $\gamma$ is a sinusoidal embedding function. The embedding function allows us to map the 2D positions to a higher dimensional Fourier space which has been shown to help in improved learning~\cite{tancik2020fourier,mildenhall2021nerf}. Note that we use a single Local-CPT to process all regional Contour-Pose components, and provide $\mathbf{l}_{r}$ after $\mathcal{F}_1$ as the region indicator. The features in $F_{r}$ are then averaged pooled with $(n+1)$ stride to $F^{p}_{r}\in\mathbb{R}^{T\times3\times C_{N}}$ to condense the point-wise dimension. Based on this design, every layer in Local-CPT has a complexity of $\mathcal{O}(\frac{n^2}{5})$ by focusing on local parts.

\begin{table}
    \centering
    \vspace{-3mm}
    \caption{The statistics of four large-scale datasets}
    \setlength{\tabcolsep}{2pt}
    \vspace{-2mm}
    \begin{tabular}{ccccc}
    \toprule
        Dataset & Id & Seq & Frames/Seq & Distractor\\
    \hline
        OUMVLP & 10,307 & 288,696 & 24.88&\xmark\\
        GREW & 26,345 & 128,671 &109.92 & \cmark\\
        Gait3D & 4,000 & 25,309 & 129.57 & \xmark\\
        SUSTech& 1050 & 25,216 & 91.61 & \xmark\\
        CCPG & 200 & 16,282 & 107.14 & \xmark \\
        BRIAR  & 1,216 & 84,913 &1943.25 & \cmark\\
        
    \bottomrule
    \end{tabular}

    \label{tab:datasets}
    \vspace{-5mm}
\end{table}

\subsubsection{Global Pose-Feature Transformer}

Once the regional features are computed, a Global Pose-Feature Transformer (Global-PFT) is used to compute a global ID representation. Similar to Local-CPT, Global-PFT is built on TTL, where each layer is denoted as $\mathcal{M}_i$. The forward process for Global-PFT can defined as follows:
\begin{gather}
    I^{g} = AvgPool(\mathcal{M}_X \circ \mathcal{M}_{X-1} \circ ... \mathcal{M}_{1}(F^{p}), V),\\
    F^{p} = F^{p}_{h} \oplus F^{p}_{la} \oplus F^{p}_{ra} \oplus F^{p}_{ll} \oplus F^{p}_{rl},
\end{gather}
where $F^{p}\in\mathbb{R}^{T\times 15\times C_{N}}$ is the concatenated feature of regional features in $F^{p}_{r}$; Global-PFT takes $F^{p}$ and to obtain the subject's identification embedding in $I^{g}\in \mathbb{R}^{1\times S_{X}}$, where $S_{i}$ is the channel size of $\mathcal{M}_i$.

\begin{table*}[!htb]
\setlength{\tabcolsep}{2pt}
\renewcommand{\arraystretch}{1.11}
\caption{Quantitative comparison of \textcolor{LimeGreen}{\textbf{keypoint-based}} gait recognition methods across six large scale datasets. The best performers are colored, and the second best methods are \underline{underlined}. GaitTR has a similar compute cost but a lower performance. GaitContour outperforms GPGait with a smaller template size and compute cost. The Ver represents TAR@FAR=$10^{-3}$. }
    \footnotesize
    \centering
    \begin{tabular}{ c |c |c|c| c| c c | c c |c c c | c c |c c }
    \hline
         \multirow{3}{*}{Method} & \multirow{3}{*}{Params(M)} & \multirow{3}{*}{FLOPs(G)} & \multirow{3}{*}{Template size}&  \multicolumn{12}{c}{Testing Datasets} \\
        \cline{5-16}
         &  &  &  & OUMVLP & \multicolumn{2}{|c|}{GREW}& \multicolumn{2}{|c|}{Gait3D} &\multicolumn{3}{|c|}{BRIAR} & \multicolumn{2}{|c}{SUSTech1K} & \multicolumn{2}{|c}{CCPG Full}\\
         \cline{5-16}
           &  & & & R1 & R1 & R5  & R1 & R5  & R1 & R5 & Ver & R1 & R5 & R1 & mAP\\
\hline
         GaitGraph2~\cite{teepe2022towards} & 0.76 & \textcolor{LimeGreen}{\textbf{0.58}} &  \underline{$1 \times 384$}& \textcolor{LimeGreen}{\textbf{62.1}} &33.5 & - & 11.1 & -  & 6.8&16.2 &1.5 & 18.6 & 40.2 &5.0 & 2.4\\
         GaitTR~\cite{zhang2023spatial} & \textcolor{LimeGreen}{\textbf{0.55}} & \underline{0.98} & \textcolor{LimeGreen}{$\mathbf{1\times 256}$} & 39.8 & 48.6 & 65.5 & 7.2 & 16.4&\underline{42.2} &\underline{68.7} &30.5 & 30.8 & 56.0 & 24.3 & 9.7 \\
        GPGait~\cite{fu2023gpgait} & 7.93 & 3.38 & $19\times 256$ & 59.1 & \underline{57.0} & \underline{68.5} & \underline{22.4} & \underline{35.9} & 38.0& 56.6 & \underline{32.0} & \underline{47.4} & \underline{70.6} & \underline{54.7} & \underline{25.8}\\
        GaitContour(Ours) & \underline{0.66} & 1.41 & \textcolor{LimeGreen}{$\mathbf{1 \times 256}$} & \underline{60.8} & \textcolor{LimeGreen}{\textbf{57.4}}  & \textcolor{LimeGreen}{\textbf{72.9}} & \textcolor{LimeGreen}{\textbf{25.3}} & \textcolor{LimeGreen}{\textbf{41.3}} &\textcolor{LimeGreen}{\textbf{55.2}} & \textcolor{LimeGreen}{\textbf{74.6}}& \textcolor{LimeGreen}{\textbf{40.0}} & \textcolor{LimeGreen}{\textbf{55.5}} & \textcolor{LimeGreen}{\textbf{72.3}} & \textcolor{LimeGreen}{\textbf{57.8}} & \textcolor{LimeGreen}{\textbf{27.1}}\\
    \hline

    \end{tabular}
    
    \label{tab:maintable}
    \vspace{-3mm}
\end{table*}

During training, triplet loss~\cite{DBLP:journals/corr/HofferA14} is applied to maximize the distance of representations from different identities and minimize the ones from the same identity.

\section{Experiments}
\label{sec:experiments}

\subsection{Datasets and Metrics}
We evaluate our method on several large scale datasets, i.e. OUMVLP~\cite{an2020performance}, CCPG~\cite{li2023depth}, SUSTech1K~\cite{shen2023lidargait}, GREW~\cite{zhu2021gait}, Gait3D~\cite{zheng2022gait} and BRIAR~\cite{cornett2023expanding}. The first three are constrained datasets and the rest are unconstrained datasets to assess the performance of challenging real-world scenarios. The statistics of these datasets are in~\Cref{tab:datasets}. One highlight of BRIAR is that its video clips are of much longer duration, at around 1900 frames per sequence, compared to around 110 frames per sequence for other gait datasets. BRIAR also has 70 sequences per identity, compared to datasets such as OUMVLP, which has 30 sequences per identity. This abundance of temporal information allows us to explore trade-offs in real applications, \eg, when image-based methods become too computationally expensive on inputs with a large number of frames.

\noindent\textbf{Evaluation Metric} For OUMVLP, Gait3D, SUSTech1K and GREW, \textit{rank retrieval} is employed to evaluate gait recognition performance. For the BRIAR dataset, we also measure performance using Receiver-Operating Characteristics (ROC) curves.
For different gait recognition algorithms, 
Additionally, we compare the number of parameters, Floating Point Operations (FLOPs), and output template size to understand their efficiency in performance and storage. Note that, we only account for the parameters and FLOPs of the encoder.
The implementation details are provided in the supplementary material section.

\subsection{Quantitative Evaluation}
As summarized in Table~\ref{tab:maintable}, we compare GaitContour against other SOTA keypoint-based gait recognition algorithms. All evaluation results other than those for BRIAR come from the respective original papers.
GaitContour performs significantly better compared to current keypoint-based methods across most datasets. In particular, GaitTR~\cite{zhang2023spatial} and GaitContour both use a Transformer-based architecture. Despite the 10X larger input size for Contour-Pose, GaitContour is similar to GaitTR~\cite{zhang2023spatial} in model size, FLOPs, and template size, while achieving much better performances across all datasets. Compared to GPGait~\cite{fu2023gpgait}, which is 12/19X larger in model/template size, GaitContour achieves better results. This demonstrates both the rich information in our novel input representation and the efficiency of processing it with GaitContour.

\subsection{Ablation Study}
\label{subsub:ablation}

\begin{table}[h]
\caption{Ablation study on region embedding (Region), a shared Local-CPT (Shared CPT), and sinusoidal embedding (Sin). The results are shown using rank retrieval, mean Average Precision(mAP) and mean Inverse
Negative Penalty (mINP) with model size. Results are based on Gait3D~\cite{zheng2022gait}.}
    \centering
    \footnotesize
\resizebox{!}{1cm}{
    \begin{tabular}{ccc |ccccc}
    \toprule
        Region & Shared CPT & Sin & Rank-1 & Rank-5 &  mAP & mINP& Param(M) \\
        \midrule
         & \cmark &  & 21.42 & 37.23  & 16.38 & 7.45 & 0.56\\
         & \cmark &\cmark & 23.32 & 40.74  &17.48 & 7.65& 0.56 \\        
         \cmark & \cmark & &22.92 & 39.24  & 17.05 & 8.30 & 0.66 \\
         \cmark &  & \cmark & 23.12 & 40.84  & 17.42 & 7.61 & 1.78 \\
         \cmark & \cmark & \cmark & 25.32 & 41.34  & 18.62 & 8.34 & 0.66 \\
         \bottomrule
    \end{tabular}}
\label{tab:Effect_parts}
\end{table}

\noindent\textbf{GaitContour architecture design.} 
\noindent In this work, we proposed three key design concepts: \textit{regional embedding}, \textit{a shared Local-CPT for different regions}, and \textit{sinusoidal embedding}. As shown in~\Cref{tab:Effect_parts}, we perform several ablation studies to confirm their usefulness. We find that regional embedding and sinusoidal embedding improve performance by 1.5\% and 1.9\% respectively; when both techniques are used together, we achieve a 3.9\% improvement overall. Furthermore, if we use five individual transformers, with the same structure as Local-CPT, the overall performance degrades by 2.2\% and the model size goes up by 1.12M parameters, 269\% of a shared Local-CPT. Using a shared transformer to process all regional information allows more augmentation on the input side and an overall more performant model.

\begin{table}
    \vspace{-3mm}
    \centering
    \setlength{\tabcolsep}{2pt}
    \small
    \caption{Comparison among different keypoint-based inputs and Contour-Pose applied on various keypoint-based methods. $\textrm{Contour-Pose}_{\textrm{NA}}$ stands for a Contour-Pose configuration with no clock-wise arrangement.}
    \resizebox{!}{1.2cm}{
    \begin{tabular}{c| c| c |c|c}
    \hline
       Index & Method & Representation &FLOPs(G) & Gait3D \\
        \hline
       (a)& GaitTR~\cite{zhang2023spatial} & Pose (17) & 0.98 & 7.2\\
        \hline
       (b) & GaitTR~\cite{zhang2023spatial} & Contour (112) & 1.81 & 4.5\\
        \hline
        (c) & GaitTR~\cite{zhang2023spatial} & Contour-Pose (165) & 2.83& 19.3\\
        \hline
        (d)& GaitGraph~\cite{teepe2021gaitgraph} & Contour-Pose (165) & 1.97 & 8.7\\
        \hline
        (e) & GaitGraph2~\cite{teepe2022towards} & Contour-Pose (165) & 8.37 & 16.2 \\
        \hline
        (f) &GaitContour & $\textrm{Contour-Pose}_{\textrm{NA}}$ (165) & 1.41 & 13.9\\
        \hline
        (g) &GaitContour & Contour-Pose (165)&1.41 & 25.3\\
        
        \hline

    \end{tabular}
    }
    \vspace{-2mm}
    \label{tab:input}
\end{table}

\begin{table}
    \centering
    \setlength{\tabcolsep}{2pt}
    \small
    \caption{Comparison under the small template size or model size.}

    \begin{subtable}[h]{0.5\textwidth}\centering
    
       \resizebox{!}{1.1cm}{ \begin{tabular}{c|cc|cc}
    \toprule
        Model & Parameters(M) & FLOPs(G) & Gait3D & GREW \\
        \midrule
        GaitGL~\cite{lin2021gait} &11.19 & 58.55 & 29.7 & 47.3 \\
        GaitBase~\cite{fan2023opengait} & 7.30 & 9.45 & 64.6 & 60.1 \\
         GaitGL$_{tiny}$& 0.77 & 1.45 & 12.2 & 17.4 \\
         GaitBase$_{tiny}$& 0.72 & 1.62 & 18.2 & 3.6 \\
         GaitContour & 0.66 & 1.41 & 25.3 & 57.4 \\
    \bottomrule
    \end{tabular}
    }
    \caption{Comparison under comparable model size}
    \end{subtable}
    \par\vskip+3pt

    \begin{subtable}[h]{0.5\textwidth} \centering
    \resizebox{!}{0.9cm}{
    \begin{tabular}{c|c|cc}
    \toprule
        Model & Template size &  Gait3D & GREW \\
        \midrule
         GaitBase~\cite{fan2023opengait}& $16 \times 256$ &  64.6 & 60.1 \\
         GaitBase$_{squeeze}$& $1 \times 256$ & 50.5 & 40.7 \\
         GaitContour & $1\times 256$ &  25.3 & 57.4 \\
    \bottomrule
    \end{tabular}
    }
    \caption{Comparison under models with the same template size}
    \end{subtable}
    
    \vspace{-7mm}
    \label{tab:template}
\end{table}

\noindent\textbf{The effectiveness of Contour-Pose and temporal consistency.} In ~\Cref{tab:input}, we explore different keypoint-based gait representations to show the effectiveness of Contour-Pose. GaitTR~\cite{zhang2023spatial} is used as the method to benchmark different inputs based on their performances on the Gait3D dataset. The vanilla GaitTR (a) achieves an improvement of 7.2\% with pose keypoints. One direct approach to leverage silhouette information is to use contour points directly; to this end, we directly sample 112 contour points to construct a directed graph, where each contour point has two edge connections to the front and back contour points. This representation (b) leads to a decrease in performance when applied to GaitTR. If we use a pose-guided sampling scheme to construct Contour-Pose instead, we see a significant accuracy improvement on GaitTR from 7.2\% to 19.3\% (c). We believe that this improvement comes from an established order based on pose keypoints across frames, i.e., temporal consistency. We further demonstrate the importance of point ordering in Contour-Pose by examining the performance of $\textrm{Contour-Pose}_{\textrm{NA}}$, where the contour points are not
 sorted in a clockwise order. This leads to a significant decrease in performance from 25.3\% to 13.9\% on GaitContour (f).

\noindent\textbf{The effect of template size.}
Template size is the subject identity embedding size produced by the gait recognition model. Larger templates can contain more information but are more expensive to store. In real applications, a vast number of identity embeddings needs to be stored and compared, making the template size a key consideration. As shown in~\Cref{tab:template}, we examine the performance of GaitBase~\cite{fan2023opengait} with an equivalent template size to GaitContour. We note that the GaitBase~\cite{fan2023opengait} backbone is unchanged; instead, only the output template size is reduced from $16\times 256$ to $1 \times 256$, noted as GaitBase$_{squeeze}$.
Despite its significantly larger model size, GaitBase's performance drops from 64.6\% to 50.5\% on Gait3D and 60.1\% to 40.7\% on GREW with a constrained template size. On the GREW dataset, GaitContour even achieves 16.7\% better performance compared to GaitBase with the same template size. This demonstrates the necessity for image-based methods to have a large template size to achieve good performance, and the efficiency of GaitContour, an effective way to concentrate sequences to a small feature size.

\noindent\textbf{The effect of model size on performance.}
The size and computational cost of a model are also crucial aspects for real deployment. Larger models are better function approximators, but are more costly to train and infer; this is a non-trivial issue particularly in gait recognition when multi-frame inputs can be very high in dimensionality.  Image-based methods generally construct larger models, as shown in Table~\ref{tab:imagetable}. If we reduce the model size for these image-based models, as shown in~\Cref{tab:template}, to be comparable to GaitContour, their performances significantly degrade. To maintain the original backbone structures, we only reduce the channel numbers. Interestingly, GaitBase's performance dropped from 60.1\% to 3.6\% on the GREW dataset, demonstrating these large-scale architectures cannot be straightforwardly reduced to achieve a balance between efficiency and performance.

\begin{figure}[h]
    \centering
    
        \begin{subfigure}[b]{.495\linewidth}
            \includegraphics[height=.75\textwidth]{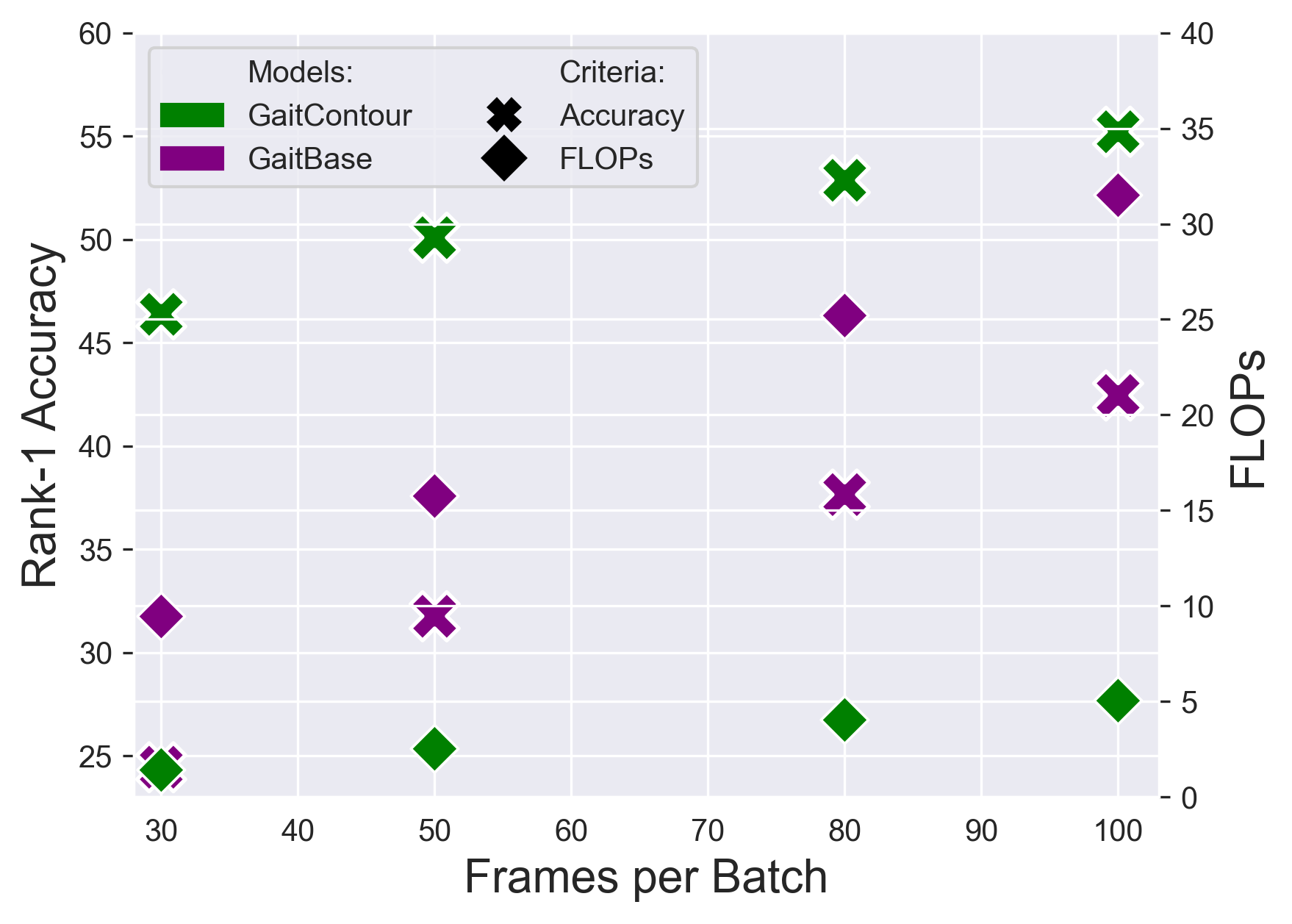}            
            \caption{\scriptsize The changes of performance and FLOPs with different \textbf{temporal windows}. GaitContour is significantly more efficient and has better performance. }
            \label{fs}
        \end{subfigure}
        \begin{subfigure}[b]{.495\linewidth}
            \includegraphics[height=.75\textwidth]{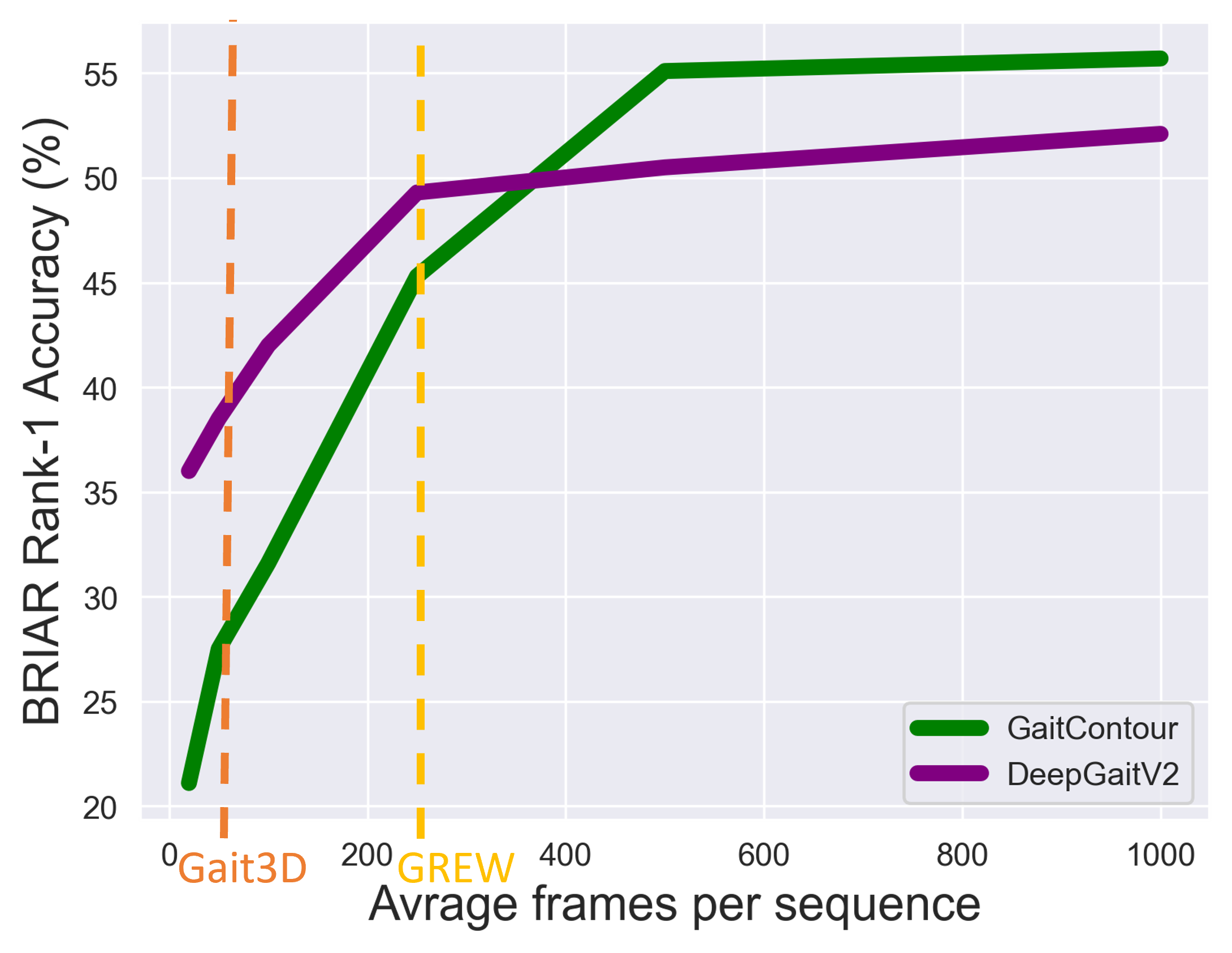}
            \caption{\scriptsize The changes of performance with different \textbf{temporal diversity}. GaitContour needs more temporal diversity during training to achieve good performance.}
            \label{fg}
        \end{subfigure} \\
    \vspace{-2mm}
    \caption{The effect of temporal information during training. Results are evaluated on the BRIAR dataset.}
   \label{fig:temporal}
   \vspace{-4mm}
\end{figure}

\begin{table*}[!htb]
\setlength{\tabcolsep}{2pt}
\renewcommand{\arraystretch}{1.11}
\caption{Quantitative comparison of \textcolor{Turquoise}{\textbf{image and fusion-based}} gait recognition methods across six large scale datasets. The best performers are colored, and the second best methods are \underline{underlined}. The Ver represents TAR@FAR=$10^{-3}$. }
    \footnotesize
    \centering
    \begin{tabular}{ c |c |c|c| c| c c | c c |c c c | c c |c c }
    \hline
         \multirow{3}{*}{Method} & \multirow{3}{*}{Params(M)} & \multirow{3}{*}{FLOPs(G)} & \multirow{3}{*}{Template size}&  \multicolumn{12}{c}{Testing Datasets} \\
        \cline{5-16}
         &  &  &  & OUMVLP & \multicolumn{2}{|c|}{GREW}& \multicolumn{2}{|c|}{Gait3D} &\multicolumn{3}{|c|}{BRIAR} & \multicolumn{2}{|c}{SUSTech1K} & \multicolumn{2}{|c}{CCPG Full}\\
         \cline{5-16}
           &  & & & R1 & R1 & R5  & R1 & R5  & R1 & R5 & Ver & R1 & R5 & R1 & mAP\\
\hline
         GaitSet~\cite{chao2019gaitset} & \underline{6.31} & 12.91 & $62\times 256$ &  87.1 & 46.3 & 63.6 & 36.7 & 58.3 &  39.4 & 60.2 & 30.5 & 65.0 & 84.8 & 77.7 &46.4\\
          GaitPart~\cite{fan2020gaitpart} & \textcolor{Turquoise}{\textbf{4.84}} & \textcolor{Turquoise}{\textbf{7.93}} & \textcolor{Turquoise}{$\mathbf{30\times 128}$} & 88.5 & 44.0 & 60.7& 28.2 & 47.6 & 41.4 &61.5 & 32.3 & 59.2 & 80.8 & 77.8 &45.5\\
          GaitBase~\cite{fan2023opengait} & 7.30 & \underline{9.45} & \underline{$16\times 256$} & \underline{90.8} & 60.1 & - & 64.6 & 74.7 &42.4 &63.5 &32.2 & 76.1 & 89.4 & - & -\\
          GaitGL~\cite{lin2021gait} & 11.19 & 58.55 & $64\times 256$ & 89.7 & 47.3 & 64.4 & 29.7 & 48.5  & 52.0&70.7 & 39.8 &63.1 & 82.8 & 69.1 & 27.0\\
         DeepGaitV2~\cite{fan2023exploring} & 13.20 & 569.00 & \underline{$16\times 256$} & \textcolor{Turquoise}{\textbf{91.9}} & 77.7 & 88.9 & 74.4 & 88.0  & 52.2&70.2 & 46.5 & 77.4 & 90.2 & \underline{90.3} &62.0\\
        BiFusion~\cite{peng2023learning} & 7.56 & 8.26 & \underline{$16\times 256$} & 89.9 & 45.5 & 64.5 & 30.8 & 49.9  & 48.5&63.4 & 36.5 & 62.1 & 83.4 & 77.5 &46.7\\
        SkeletonGait++~\cite{fan2023skeletongait} & 13.27 & 91.79 & \underline{$16\times 256$} & - & \underline{85.8} & \underline{92.6} & \underline{77.6} & \underline{89.4}  & \underline{57.9} & \underline{76.6} & \underline{51.5} & \underline{81.3} & \underline{95.5} & 90.1 & \underline{63.6}\\
        Contour-Pose++& 13.27 & 91.79 & \underline{$16\times 256$} & - & \textcolor{Turquoise}{\textbf{86.1}} & \textcolor{Turquoise}{\textbf{93.4}} & \textcolor{Turquoise}{\textbf{79.6}} & \textcolor{Turquoise}{\textbf{89.8}}  & \textcolor{Turquoise}{\textbf{59.7}} & \textcolor{Turquoise}{\textbf{77.6}} & \textcolor{Turquoise}{\textbf{53.1}} & \textcolor{Turquoise}{\textbf{83.3}} & \textcolor{Turquoise}{\textbf{95.8}} & \textcolor{Turquoise}{\textbf{92.1}} & \textcolor{Turquoise}{\textbf{67.1}}\\

\hline

    \end{tabular}
    
    \label{tab:imagetable}
    \vspace{-5mm}
\end{table*}

\subsection{Fusion-based Comparison with Contour-Pose} 

We further analyze the utility of Contour-Pose in fusion-based gait recognition models. Fusion-based gait recognition~\cite{fan2023skeletongait, peng2023learning} leverages both silhouette and key-point representations to improve recognition performances. To this end, we propose a fusion method called Contour-Pose++, which uses a large backbone model in similar fashion to SkeletonGait++~\cite{fan2023skeletongait}. Specifically, we map Contour-Pose to a 2D skeletonmap, extracting features from the skeletonmap and silhouette image using two CNN-based networks. These features are fused in a large backbone, i.e. DeepGaitV2. Please refer to the supplemental material for more specific architecture design.

As shown in~\Cref{tab:imagetable}, Contour-Pose++ consistently outperforms other image or fusion-based methods on five benchmark datasets. Notably, even though SkeletonGait++ includes a silhouette branch incorporating body shape information with pose keypoints, Contour-Pose++ yields significant improvements, e.g., 3\% and 2\% on Gait3D and SUSTech1.  This demonstrates the effectiveness of Contour-Pose in encoding discriminative gait features. 

\subsection{Analysis on Temporal Information}

In principle, gait recognition explores temporal patterns to perform biometrics; however, this aspect of gait recognition has seldom been analyzed due to the frame number limitation in popular datasets. As we demonstrate in~\cref{fs}, for image-based and keypoint-based methods, models trained with a larger temporal window size can obtain better performance. This is particularly useful if the dataset has long-duration sequences, \eg, in BRIAR. Training with more frames requires more computation, especially when the model itself is already large. As we showed in~\cref{fs}, large models like GaitBase not only have more FLOPs than GaitContour, but a steeper rate at which their FLOPs increase given more frames. In fact, GaitContour uses fewer FLOPs with a 100-frame input compared to a 30-frame to GaitBase. The performance gap between GaitContour and GaitBase also increases as the temporal window size goes down, likely because GaitBase overfits more on smaller details due to its dense inputs and large model. 

From~\Cref{tab:maintable} and~\Cref{tab:imagetable}, we observe a smaller performance gap between keypoint-based and image-based methods on the BRIAR dataset compared to other datasets. This is a key difference between the BRIAR dataset and other benchmark sets with few frames. The abundance of temporal information in the BRIAR dataset enhances keypoint-based method performance. Keypoint-based methods can only observe compressed spatial information, making it challenging to extract discriminative subject features. Therefore, more temporal diversity is required during the training phase to build robust subject features. The substantial volume of temporal information in the BRIAR dataset allows keypoint-based methods to extract finer differences between subjects, achieving even higher performance than some image-based methods. The performance gap observed in GREW and Gait3D further supports this assumption. We conduct experiments on the BRIAR dataset to validate this finding. As shown in~\cref{fg}, when less temporal diversity is employed during training, the performance of GaitContour drops more significantly than DeepGaitV2, demonstrating that the keypoint-based methods need more temporal diversity during training. Previous public datasets, such as GREW, have limited temporal diversity—only 13.3\% of that in BRIAR—leading to the lower performance of keypoint-based methods.

\subsection{Discussion}

\noindent\textbf{Limitation and Prospect} Regarding the potential drawbacks of Contour-Pose, occlusion and limited field of view can result in imperfect silhouettes, particularly in uncontrolled environments. Such imperfections can pose challenges during training, as boundaries in adjacent frames may vary significantly. Currently, we apply point-wise augmentation to mitigate this problem, as we did for silhouettes. 
We believe GaitContour still holds great potential. While there remains a performance gap between GaitContour and image-based methods, GaitContour demonstrates effective performance when there is abundant temporal diversity during training. Furthermore, due to its lightweight and efficient nature, GaitContour can serve as a plug-in module for any image-based method. Based on previous works~\cite{cui2023multi,peng2023learning}, the combination of keypoint-based and image-based methods consistently improves the overall performance.

\section{Conclusion}
\label{sec:conclusion}

In this work, we propose a novel gait representation called Contour-Pose and a gait recognition model, GaitContour, that leverages the advantages of Contour-Pose to achieve significant improvements in performance and efficiency. Contour-Pose uses a pose-guided sampling process on a silhouette, which approximates contour points from silhouette edges and sample points based on distances from pose keypoints. This representation efficiently preserves information from both silhouette and pose keypoints, and is temporally consistent. We can observe significant performance improvements when Contour-Pose is applied to various keypoint-based recognition models. We further develop GaitContour, which is tailored to analyze Contour-Pose. GaitContour contains two components: Local-CPT, and Gloabl-PFT. Local-CPT analyzes Contour-Pose at five different local regions and aggregates the outputs to a sparse global feature. Global-PFT then generates a subject identity embedding based on this global feature. Compared to a conventional Transformer, this local-to-global design significantly improves the model efficiency. Our experiments show that GaitContour is on par with SOTA image-based methods on a practical dataset, while maintaining efficiency in model size, template size, and FLOPs to that of keypoint-based methods, thereby making gait recognition much more practical for real applications.

\section{Acknowledgement}

This research is based upon work supported in part by
the Office of the Director of National Intelligence (ODNI),
Intelligence Advanced Research Projects Activity (IARPA),
via [2022-21102100005]. The views and conclusions contained
herein are those of the authors and should not be
interpreted as necessarily representing the official policies,
either expressed or implied, of ODNI, IARPA, or the U.S.
Government. The US. Government is authorized to reproduce
and distribute reprints for governmental purposes notwithstanding
any copyright annotation therein.

\newpage
{
    \bibliographystyle{splncs04}
    \bibliography{main}
}

\end{document}

\maketitle

\appendix

\setcounter{page}{1}

\renewcommand{\thefigure}{\Alph{figure}}
\renewcommand{\thetable}{\Alph{table}}
\setcounter{equation}{9}
\setcounter{table}{0}
\setcounter{figure}{0}

We provide additional
details on key sections as described below:
\begin{enumerate}
    \item \textbf{Contour-Pose}: Details of the Contour-Pose construction process~(\ref{sec:construction process}), visualization~(\ref{sec:visualzation}) and failure cases~(\ref{sec:fail}).
    \item \textbf{GaitContour}: Sinusoidal embedding~(\ref{sec:sinusoidal embedding}) and the temporal transformer layer~(\ref{sec:TTL}).
    \item  \textbf{Experimental Details}: BRIAR dataset~(\ref{briar}), BRIAR visualizations~(\ref{sec:challenge}) and implementation details~(\ref{implementation}).
    \item \textbf{Additional Ablation Study}: Ablation on the number of contour points (\ref{numofpoints}), network latency and memory (\ref{latency}), encoding methods (\ref{encoding}), body parts (\ref{bodypart}), fusion (\ref{fusion}), cross-domain evaluation (\ref{cross}) and Contour-Pose++ architecture (\ref{++})
\end{enumerate}

\section{Contour-Pose}

\subsection{Construction Process}
\label{sec:construction process}

The data format of pose and silhouette is often inconsistent across different datasets. In practice, silhouettes are usually preprocessed through a set of conventional normalization processes~\cite {chao2019gaitset}, such that the final frame sequences share the same center and the same size. Poses are usually provided based on coordinates from the original frames. To align these two representations, we chose to center and scale the pose as follows:

\begin{equation}
\begin{split}
    x^{aligned}_v = \frac{x_v - x_{center}}{y_{max}-y_{min}} \times H' + \frac{1}{2} \times W,\\
    y^{aligned}_v = \frac{y_v - y_{center}}{y_{max}-y_{min}} \times H' + \frac{1}{2} \times H,
\end{split}
\end{equation}
where $(x_{center}$,$y_{center})$ represents the coordinates of the centroid of the human body, and $y_{max}$ and $y_{min}$ correspond to the extreme value on the y-axis among $V$ nodes. Additionally, $H'$ denotes the unified height among poses which is slightly smaller than $H$ to appropriately adjust the normalized pose to fit within the silhouette. After the above operations, the pose and silhouette share a common center, i.e. $(\frac{1}{2} W, \frac{1}{2}H)$. Furthermore, this step filters out unrelated information, such as the trajectory and the frame size. 
We then extract contour points from the silhouettes. Inspired by the shape model~\cite{baumberg1994learning}, we employ a Teh-Chin algorithm~\cite{teh1989detection} to sample points on silhouettes $\mS$. For each pose keypoint, 
the $N$ nearest contour points are selected. After that, we uniformly pick $M$ points from them, e.g., one out of every $N/M$ points in $N$ points which are sorted based on their distance to the pose keypoint, which prevents the oversampling in small detailed areas. The process is summarized in~\Cref{alg:BuildCP}. The $\texttt{clockwise}$ operation is accomplished by computing the angles between the selected contour points and associated keypoints, and subsequently sorting the contour points based on these angles in ascending order. 

In this work, we define the centroid of the human body as follows:  
\begin{equation}
\begin{split}
    x_{center} &= \frac{x_{la}+ x_{ra} + x_{ll} + x_{rl} }{4}, \\
    y_{center} &= \frac{y_{max} + y_{min}}{2},
\end{split}
\end{equation}
where $x_{center}$ uses the average of all limbs to mitigate the noise introduced by pose estimation and $y_{center}$ is the middle point of the body. The normalized height $H'$ is set as 110 when silhouette with the square size 128. For each skeleton point, we first select $N=40$ points and then get the sparsest $M=10$ points to capture the body shape information.

\subsection{Visualization}
\label{sec:visualzation}

As shown in~\cref{fig:visualization}, the proposed Contour-Pose captures the body shape and pose information compactly in various datasets. Especially, the silhouette is ambiguous when self-occlusion appears, but Contour-Pose helps the model to learn internal structure better.

\begin{algorithm}[!t]
\footnotesize
\caption{\label{alg:BuildCP} Building Contour-pose}
\begin{algorithmic}
    \STATE \textbf{Input:} Silhouette sequence $\{\vs_t\}$ Aligned poses sequence $ \{\vp_t \in \mathbb{R}^{V \times 2}\},t\in[1,\cdots,T]$, The number of nearest points $N$, The number of contour point per keypoint $M$
    \STATE \textbf{Output:} Contour-Pose $\mC\mP = \{\vc\vp_t\}_{t=1}^{T}$, 
    \STATE \textbf{for} frame $t = 1, \ldots, T$ \textbf{do}
        \STATE $~~~~$ \textcolor{gray}{\# Get contour points of silhouette }
        \STATE $~~~~$ $\vc_t=\texttt{findContours}(\vs_t)$ 
        \STATE $~~~~$ \textbf{for} keypoint $v = 1, \ldots, V$ \textbf{do}
            \STATE $~~~~$ $~~~~$ \textcolor{gray}{\# Get the distance of contour points to a keypoint}
            \STATE $~~~~$ $~~~~$ $\vd_{t,v} = \texttt{Euclidean}(\vp_{t,v}, \vc_t) $
            \STATE $~~~~$ $~~~~$ \textcolor{gray}{\# Sort the contour point based on the distance to the point}
            \STATE $~~~~$ $~~~~$ $\vc_t = \texttt{sort}(\vc_t, key = \vd_{t,v})$
            \STATE $~~~~$ $~~~~$ \textcolor{gray}{\# Get the Contour-Pose}
            \STATE $~~~~$ $~~~~$ $\vc\vp_{t,v} = \vc_t[:N:N/M]$
            \STATE $~~~~$ $~~~~$ \textcolor{gray}{\# Sort Contour-Pose to clockwise}
            \STATE $~~~~$ $~~~~$ $\vc\vp_{t,v} = \texttt{clockwise}(\vc\vp_{t,v})$
            \STATE $~~~~$ $~~~~$ \textcolor{gray}{\# Remove selected Contour-Pose point from contour points to prevent the same points being sampled again}
            \STATE $~~~~$ $~~~~$ $\vc_t = \texttt{delete}(\vc_t, \vc\vp_{t,v})$
        \STATE $~~~~$\textbf{end for}
    \STATE \textbf{end for}
    \STATE return $\{\vc\vp_t\}_{t=1}^{T}$
\end{algorithmic}
\end{algorithm}

\subsection{Failure Cases}
\label{sec:fail}

\begin{figure}
\centering
\setlength{\tabcolsep}{-1pt}

\begin{tabu}{cccc}
\begin{subfigure}{0.08\textwidth}
    \includegraphics[width=\textwidth]{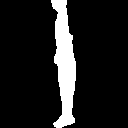}
\end{subfigure} & 
\multirow{2}{*}{
\begin{subfigure}{0.16\textwidth}
    \includegraphics[width=\textwidth, height=\textwidth]{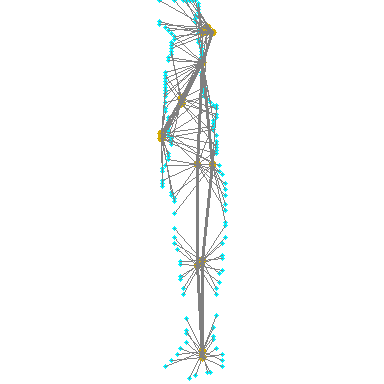}
    \end{subfigure}} &
\begin{subfigure}{0.08\textwidth}
    \includegraphics[width=\textwidth]{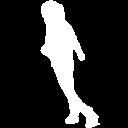}
\end{subfigure} & 
\multirow{2}{*}{
\begin{subfigure}{0.16\textwidth}
    \includegraphics[width=\textwidth, height=\textwidth]{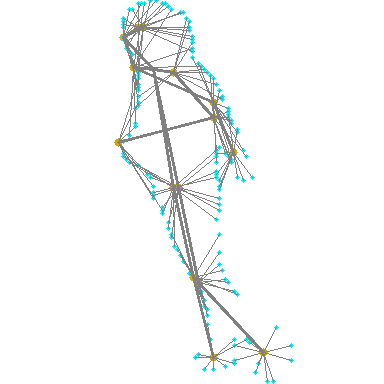}
    \end{subfigure}}
    \\ 

\begin{subfigure}{0.08\textwidth}
    \includegraphics[width=\textwidth, height=\textwidth]{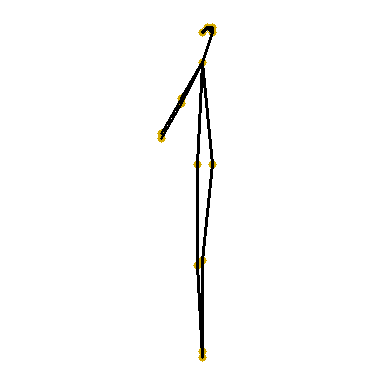}
\end{subfigure}  & & 
\begin{subfigure}{0.08\textwidth}
    \includegraphics[width=\textwidth, height=\textwidth]{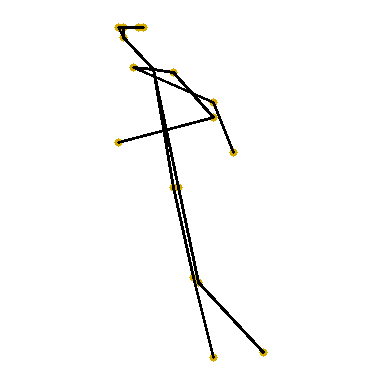}
\end{subfigure}  &\\
\end{tabu}

\caption{Contour-Pose failure cases.}
\label{fig:failure}
\end{figure}

We have also observed instances where the Contour-Pose may be noisy when the pose and silhouette cannot align effectively. This occurrence can be attributed to the following factors:
(1) \textbf{Incomplete silhouette}: The Contour-Pose relies on the assumption that both the silhouette and pose have a complete body for alignment so it is challenging to obtain a reasonable Contour-Pose with partial body silhouettes, as demonstrated in~\cref{fig:failure} on the left. (2) \textbf{Inaccurate Pose Estimation}: Inaccuracies in pose estimation can contribute to unusual body structures and subsequently impact the Contour-Pose construction. This is evident in~\cref{fig:failure} on the right, where incorrect pose estimation leads to distorted Contour-Pose representations. 

We notice that these failure cases appear across multiple frames in Gait3D and OUMVLP datasets, which may have led to diminished performance on these datasets. However, we also note that this misalignment issue is a practical issue and would not happen if we have access to the original RGB frames, which allow for direct alignment of pose and contour on the original frame. Additionally, a more mature pose estimation approach can be employed to obtain reliable keypoint locations. BRIAR is an example to demonstrate that under these assumptions which can be satisfied in real-life applications, GaitContour can achieve performance even surpassing the image-based approaches.

\section{GaitContour}

\subsection{Sinusoidal Embedding}
\label{sec:sinusoidal embedding}

We apply sinusoidal embedding to map the location information in the 2D coordinate to a higher dimensional Fourier space, which has been shown to lead to better optimization compared to using raw coordinates~\cite{mildenhall2021nerf,tancik2020fourier}. Specifically, we apply $\gamma$ to map the coordinates from $\mathbb{R}$ to $\mathbb{R}^{2L}$ separately

\begin{equation}
\begin{split}
    \gamma(x) = (\sin(2^0\pi x), \cos(2^0\pi x), \ldots, \\
    \sin(2^{L-1}\pi x),\cos(2^{L-1}\pi x))
\end{split}
\end{equation}
where $L$ represents the number of frequencies applied. We further conduct experiments to evaluate the effect of frequency level on the recognition accuracy. As shown in~\Cref{tab:frequencies}, larger $L$ applied leads to better overall performance but more parameters involved. We finally choose $L=2$ in consideration of the balance between performance and efficiency.

\begin{table}
    \setlength{\tabcolsep}{3pt}
    \centering
        \caption{Ablation study on the number of frequencies in sinusoidal embedding. Results are based on Gait3D~\cite{zheng2022gait}.}
    \begin{tabular*}{.78\linewidth}{c |ccccc}
    \toprule
         $L$ & Rank-1 & Rank-5 &  mAP & mINP& Param(M) \\
        \midrule
          1 & 24.42 & 41.14  & 18.43 & 8.34 & 0.64\\
         2 & 25.32 & 41.34  &18.62 & 8.34& 0.66 \\        
          3 & 25.42 & 43.04  & 18.93 & 9.12 & 0.71 \\
         \bottomrule
    \end{tabular*}

    \label{tab:frequencies}
\end{table}

\subsection{Temporal Transformer Layer}
\label{sec:TTL}
In Section 3.2.1, we introduce the Temporal Transformer Layer, employing Temporal Aggregation (TA) and a self-attention mechanism to capture the spatiotemporal relations between joints. Following~\cite{zhang2023spatial, plizzari2021spatial}, The TA module employs convolution along the time dimension to aggregate information. The forward process of TA is described as follows:
\begin{equation}
\begin{aligned}
& \text{TA}(*) = \text{BN}(Mish(\text{Conv}(Dropout(*))))
\end{aligned}
\end{equation}
\noindent in which $Mish$~\cite{misra2019mish} is the activation function. $\text{MHA}$ in Equation (3) is a standard self-attention~\cite{vaswani2017attention}. The feature space is constituted by a weighted sum of the node values $v$, where the weight is determined by the pairwise similarity between two nodes' corresponding query $q_{i,n}$ and key $k_{i,n}$. Subsequently, we independently conduct the previously mentioned operation  $I$ times in parallel and project their concatenated outputs, yielding the vector $\textbf{z}$. 

\begin{equation}
\begin{aligned}
    & [q_{i,n},k_{i,n},v_{i,n}] = x_nW_i^{qkv},\\
    & head_i = \sum_n softmax(\frac{q_{i,n}k_{i,n}^T}{\sqrt{d_k}})v_{i,n}, \\
    & \textbf{z} = Concat(head_1, ...,head_i,..., head_I)W^O,  \\ 
\end{aligned}
\end{equation}

\begin{table}
    \centering
    \caption{Training details for datasets. The batch size ($p$,$c$) indicates that there are $c$ clips for each person and $p$ people in a batch.}
    \begin{tabular}{cccc}
    \hline
        Dataset & Batch Size & Frames & Steps\\
    \hline
        OUMVLP & (32,16) & 60 & 150k\\
        GREW & (32, 8) & 60 & 150k\\
        Gait3D & (32,4) & 60 & 60k\\
        SUSTech1K & (8, 8) & 60 & 40k \\
        CCPG & (4,32) & 60 & 40k \\
        BRIAR  & (16,8) & 100 & 150k\\
    \hline
    \end{tabular}
    
    \label{tab:traindetail}
\end{table}

\section{Experimental Details}
\subsection{Detailed BRIAR dataset}
\label{briar}
The BRIAR~\cite{cornett2023expanding} dataset consists of 1,216 subjects. Each identity is represented by both indoor (\textit{controlled}) and outdoor (\textit{field}) sequences. In the \textit{controlled} set, structural and random walking are captured using 10 cameras positioned at different viewing angles. The \textit{field} set involves sequences collected at various distances (up to 1000m) and altitudes (by UAVs). To avoid clothing-based cues, two garment settings are applied. Additionally, multiple walking types are involved, such as holding a box, taking a cell phone, and wearing a backpack, to test the model's robustness. At test time, the \textit{controlled} and \textit{field} clips serve as gallery and probe respectively with different garment settings. This setting aims to simulate gait recognition in real-world conditions. 

\subsection{BRIAR visualizations}
\label{sec:challenge}
\begin{figure}[!tb]
    \setlength{\abovecaptionskip}{3pt}
    \setlength{\tabcolsep}{0pt}
    \renewcommand\arraystretch{0}
    \setlength{\belowcaptionskip}{-0.2cm}
    \centering
    \small
    \begin{tabular}[c]{cccccc}
    
        LD & Cell & BP & Box & Indoor & CC \\
        \begin{subfigure}[b]{.165\linewidth}
             \includegraphics[width=\textwidth,height=1.5\textwidth]{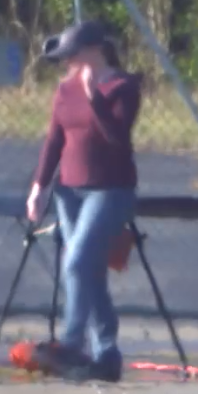}
        \end{subfigure}&
            \begin{subfigure}[b]{.165\linewidth}
             \includegraphics[width=\textwidth,height=1.5\textwidth]{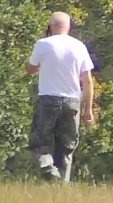}
        \end{subfigure}&
                \begin{subfigure}[b]{.165\linewidth}
             \includegraphics[width=\textwidth,height=1.5\textwidth]{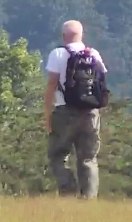}
        \end{subfigure}&
                \begin{subfigure}[b]{.165\linewidth}
             \includegraphics[width=\textwidth,height=1.5\textwidth]{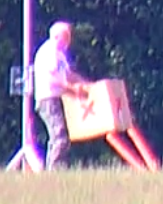}
        \end{subfigure}&
                \begin{subfigure}[b]{.165\linewidth}
             \includegraphics[width=\textwidth,height=1.5\textwidth]{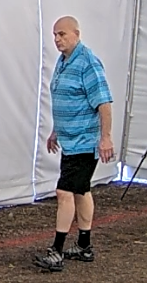}
        \end{subfigure}&
                \begin{subfigure}[b]{.165\linewidth}
             \includegraphics[width=\textwidth,height=1.5\textwidth]{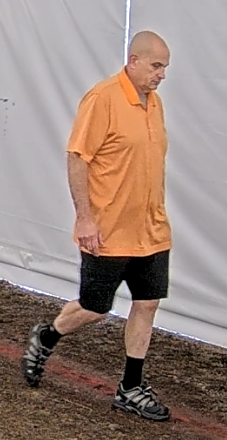}
        \end{subfigure}\\

        \begin{subfigure}[b]{.165\linewidth}
             \includegraphics[width=\textwidth,height=1.5\textwidth]{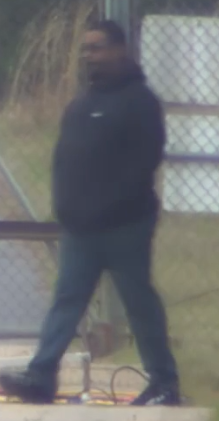}
        \end{subfigure}&
            \begin{subfigure}[b]{.165\linewidth}
             \includegraphics[width=\textwidth,height=1.5\textwidth]{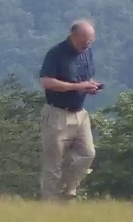}
        \end{subfigure}&
                \begin{subfigure}[b]{.165\linewidth}
             \includegraphics[width=\textwidth,height=1.5\textwidth]{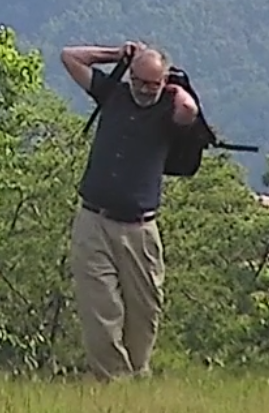}
        \end{subfigure}&
                \begin{subfigure}[b]{.165\linewidth}
             \includegraphics[width=\textwidth,height=1.5\textwidth]{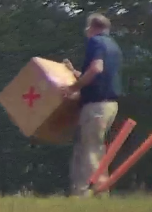}
        \end{subfigure}&
                \begin{subfigure}[b]{.165\linewidth}
             \includegraphics[width=\textwidth,height=1.5\textwidth]{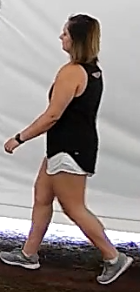}
        \end{subfigure}&
                \begin{subfigure}[b]{.165\linewidth}
             \includegraphics[width=\textwidth,height=1.5\textwidth]{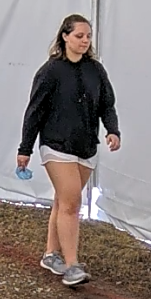}
        \end{subfigure}\\

        \begin{subfigure}[b]{.165\linewidth}
             \includegraphics[width=\textwidth,height=1.5\textwidth]{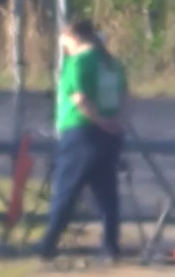}
        \end{subfigure}&
            \begin{subfigure}[b]{.165\linewidth}
             \includegraphics[width=\textwidth,height=1.5\textwidth]{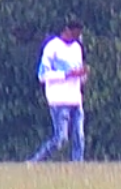}
        \end{subfigure}&
                \begin{subfigure}[b]{.165\linewidth}
             \includegraphics[width=\textwidth,height=1.5\textwidth]{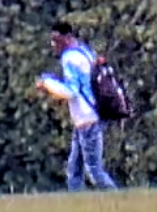}
        \end{subfigure}&
                \begin{subfigure}[b]{.165\linewidth}
             \includegraphics[width=\textwidth,height=1.5\textwidth]{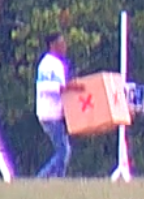}
        \end{subfigure}&
                \begin{subfigure}[b]{.165\linewidth}
             \includegraphics[width=\textwidth,height=1.5\textwidth]{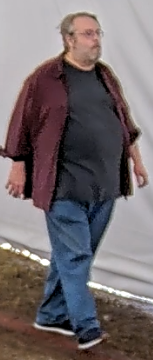}
        \end{subfigure}&
                \begin{subfigure}[b]{.165\linewidth}
             \includegraphics[width=\textwidth,height=1.5\textwidth]{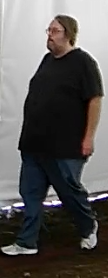}
        \end{subfigure}\\

        \begin{subfigure}[b]{.165\linewidth}
             \includegraphics[width=\textwidth,height=1.5\textwidth]{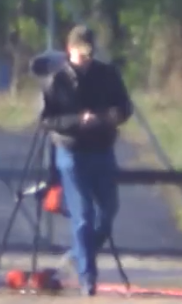}
        \end{subfigure}&
            \begin{subfigure}[b]{.165\linewidth}
             \includegraphics[width=\textwidth,height=1.5\textwidth]{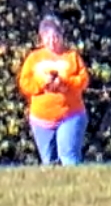}
        \end{subfigure}&
                \begin{subfigure}[b]{.165\linewidth}
             \includegraphics[width=\textwidth,height=1.5\textwidth]{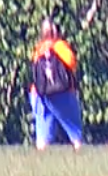}
        \end{subfigure}&
                \begin{subfigure}[b]{.165\linewidth}
             \includegraphics[width=\textwidth,height=1.5\textwidth]{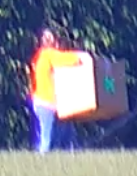}
        \end{subfigure}&
                \begin{subfigure}[b]{.165\linewidth}
             \includegraphics[width=\textwidth,height=1.5\textwidth]{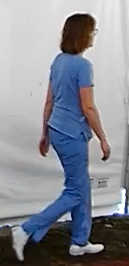}
        \end{subfigure}&
                \begin{subfigure}[b]{.165\linewidth}
             \includegraphics[width=\textwidth,height=1.5\textwidth]{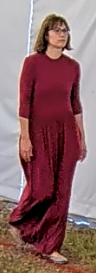}
        \end{subfigure}\\

        \begin{subfigure}[b]{.165\linewidth}
             \includegraphics[width=\textwidth,height=1.5\textwidth]{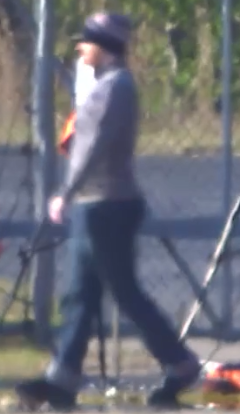}
        \end{subfigure}&
            \begin{subfigure}[b]{.165\linewidth}
             \includegraphics[width=\textwidth,height=1.5\textwidth]{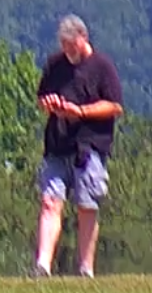}
        \end{subfigure}&
                \begin{subfigure}[b]{.165\linewidth}
             \includegraphics[width=\textwidth,height=1.5\textwidth]{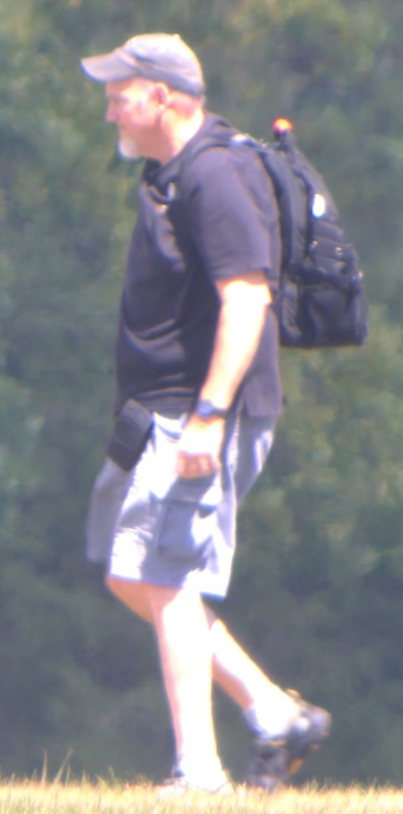}
        \end{subfigure}&
                \begin{subfigure}[b]{.165\linewidth}
             \includegraphics[width=\textwidth,height=1.5\textwidth]{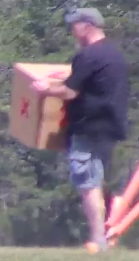}
        \end{subfigure}&
                \begin{subfigure}[b]{.165\linewidth}
             \includegraphics[width=\textwidth,height=1.5\textwidth]{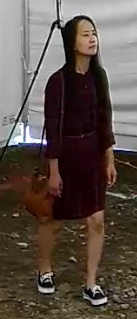}
        \end{subfigure}&
                \begin{subfigure}[b]{.165\linewidth}
             \includegraphics[width=\textwidth,height=1.5\textwidth]{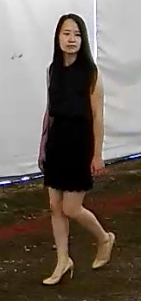}
        \end{subfigure}\\

    \end{tabular}
    \caption{Examples of subjects in different conditions from the BRIAR dataset at different walking conditions and clothing. The columns represent different challenges different to other datasets, namely Long Distance (LD, $1000m$), taking cell phone (Cell), wearing a backpack (BP), holding a box (Box), indoor walking (Indoor) and cloth changing (CC) respectively. All the subjects shown have consented to the publication of their images.}
    \label{fig:challenge}
    \vspace{-4mm}
\end{figure}

We provide some BRIAR examples in~\cref{fig:challenge}. Compared to other datasets, BRIAR takes long distances (atmosphere turbulence, low resolution), walking conditions, places, clothes and accompanying challenges, such as occlusion, into consideration, mimicking the gait recognition applied in real life. To have an intuitive feel about the dataset, we show some example clips in the folder namely \url{./clips/}. In the folder, we show the video recorded in \textbf{different platforms}, i.e.\url{./clips/controlled_rand.mp4}, \url{./clips/100m_set1.mp4} and \url{./clips/uav.mp4}, \textbf{various distances}, i.e.\url{./clips/close_high.mp4}, \url{./clips/100m_set1.mp4}, \url{./clips/400m.mp4}, and \url{./clips/500m.mp4}, \textbf{different altitudes}, i.e. \url{./clips/close_high.mp4}, \url{./clips/uav.mp4} and \url{./clips/100m_set1.mp4}, \textbf{clothes changing}, i.e. \url{./clips/100m_set1.mp4} and \url{./clips/100m_set2_stand.mp4} and \textbf{diverse movement types}, i.e. \url{./clips/100m_set2_stand.mp4} \url{./clips/controlled_struct.mp4} and \url{./clips/controlled_rand.mp4}. 
It is noticeable that \url{./clips/100m_set2.mp4} is an example of an 
  incomplete body due to vegetation occlusion. All the subjects shown have consented to the publication of their images.

\subsection{Implementation Details}
\label{implementation}

All experiments are implemented in Pytorch~\cite{paszke2019pytorch} with four NVIDIA A5000 GPUs. 
Before Contour-Pose is provided to GaitContour, we follow~\cite{zhang2023spatial} and augment the input by adding Gaussian noise with a standard deviation of 0.25 to the coordinates with 0.3 probability and horizontal flipping with a probability of 0.01. 
We add five pieces of information such as relative distance to the nose keypoint, edge distance, velocity across frames, \textit{etc}. This increases the input channel number from 2 to $5\times 2=10$.
For sinusoidal embedding, two frequency bands are used to embed these 10 features, making the final input channel number 40. The intermediate features produced by Local-CPT and Global-PFT have channel sizes of (64,64,128) and (256), respectively. We note that while traditionally the head has 5 keypoints, to reduce the sampling density at a small region, we use only 3 keypoints from the head: the left and right ear and the nose; therefore, the number of joints $V=15$ in this work. Considering the efficiency and effectiveness, $n=10$ for surrounding contour points associated with a keypoint. Adam optimizer is used with the same hyperparameters as~\cite{fu2023gpgait}. 
As shown in~\Cref{tab:traindetail}, we follow the training setting proposed in~\cite{fu2023gpgait} to have consistent comparisons. The evaluation process is conducted on $\texttt{FastPoseGait}$~\cite{meng2023fastposegait}\footnote{https://github.com/BNU-IVC/FastPoseGait}. We further include FLOPs, the number of parameters, and template size in metrics, which are based on the configuration on the GREW~\cite{zhu2021gait} dataset provided by $\texttt{OpenGait}$~\cite{fan2023opengait}\footnote{https://github.com/ShiqiYu/OpenGait}, since its scale is close to the real-life scenario. Notably, we only account for the parameters and FLOPs of the encoder getting rid of the large amount of parameters involved in the classification head during the training process. The efficiency descriptors are obtained through $\texttt{fvcore}$\footnote{https://github.com/facebookresearch/fvcore}.

\section{Ablation study}

\subsection{The number of contour points}
\label{numofpoints}

The selection of contour points involves a tradeoff between \textbf{information preservation} and \textbf{compactness}. While selecting more points introduces additional information, it also increases latency. With ten points, the area Contour-Pose reconstructed reaches \underline{97.7\%} \textbf{IoU} with the original silhouette. If we take twenty contour points, the latency will increase by 58.4\%. Furthermore, the extracted contours only have around 300 points due to low silhouette resolution, and more sampling mostly leads to redundancy.

\begin{table}
\vspace{-6mm}
    \centering
    \caption{The Latency and recognition accuracy change on Gait3D when different contour points are selected.}
    \begin{tabular}{c|c|c|c|c}
    \hline
        \# Points & 1 & 2 & 5 & 10\\
        \hline
        Gait3D Accuracy(\%) & 14.3 & 16.8 & 23.8 & 25.3\\
        Latency (ms) & 32.7 & 33.1 & 33.6 & 34.9\\
        IoU (\%) &68.8 & 76.6& 89.1& 97.7\\
        \hline
    \end{tabular}
    
    \vspace{-6mm}
    \label{tab:numpoint}
\end{table}

\subsection{Network latency and GPU memory}
\label{latency}

We also evaluate the latency of the model using an A5000 GPU.  The latency is calculated by the mean of inferring 7000 sequences with 100 frames each. Compared to other models (even point-based GPGait), GaitContour is at least 3X as fast, with building Contour-Pose incurring minimal overhead. 
GaitContour demonstrates a small GPU memory occupied, which allows us to include more sequences for a better triplet loss computation. Generally, the overall performance improves when more sequences are involved in calculating triplet loss.

\begin{table}[h]
\vspace{-3mm}
\setlength{\tabcolsep}{2pt}
\scriptsize
    \centering
    \caption{The latency and the occupied memory on a single A5000 GPU of different gait recognition backbones. GaitContour occupies less memory and takes less time, but reaches competitive gait recognition performance}
    \scalebox{0.7}{\begin{tabular}{c|c|c|c|c|c|c c}
        \hline
        \multirow{2}{*}{Models} & \multirow{2}{*}{GaitGL~\cite{lin2021gait}} & \multirow{2}{*}{GaitBase~\cite{fan2023opengait}}  & \multirow{2}{*}{DeepGaitv2~\cite{fan2023exploring}}  & \multirow{2}{*}{GPGait~\cite{fu2023gpgait}} & \multirow{2}{*}{GPGait$_{tiny}$}&  Ours  & Ours
        \\
        & & & & & &(Overall) & (Network Only)\\
        \hline
        GREW $\uparrow$ &  47.3 & 60.1 & 77.4  & 57.0 & 28.8 & \multicolumn{2}{c}{57.4} \\
        Time (ms) $\downarrow$ & 85.8 & 63.4  & 107.1 & 61.5 & 52.1 & 24.8 & 22.6\\
        Memory (GB) $\downarrow$ & 27.8 & 42.7  & 123.4& 20.5 & 14.6 & \multicolumn{2}{c}{14.1} \\
        \hline
    \end{tabular}}
    
    \vspace{-4mm}
\end{table}

\subsection{Comparison with different encoding methods}
\label{encoding}

In GaitContour, we employ a sinusoidal embedding function to map the 2D coordinates to higher frequency space, a technique that has demonstrated superiority in previous works~\cite{tancik2020fourier,mildenhall2021nerf}. Fourier transform is another technique used to transform signals to frequency space. Therefore, we examine the performance variance when employing different encoding methods.

\begin{table}
\caption{The performance on GREW dataset with different encoding methods}
    \centering
    \begin{tabular}{c| c | c}
    \toprule
        Methods & Rank-1 & Rank-5\\
    \midrule
        None & 48.1 & 66.7 \\
        Fourier & 27.5 & 44.8\\
        Sinusoidal & 57.4 & 72.9\\
    \bottomrule
    \end{tabular}
    
    \label{tab:encoding}
\end{table}

As shown in~\Cref{tab:encoding}, sinusoidal embedding achieves higher performance compared to the other method, suggesting that although the Fourier transform serves a similar function, sinusoidal embedding is a more suitable transform for points.

\subsection{The effect of different body parts as input}
\label{bodypart}
Intuitively, only certain body parts, e.g. legs, offer distinguishable information about human walking. Reducing the number of points is also essential to enhance the model's efficiency. Considering the above reasons, we conducted the experiments to identify redundant keypoints and subsequently remove them. Our definition of body parts includes the whole body, torso, and legs.

\begin{table}
    \centering
    \caption{The performance on Gait3D with different body parts as input.}
    \begin{tabular}{c|c c}
    \toprule
        Body Part & Rank-1 & Rank-5\\
    \midrule
        Whole &  25.3 & 41.3 \\
        Torso & 22.0 & 38.8\\
        Legs & 14.4 & 28.9\\
    \bottomrule
    \end{tabular}
    
    \label{tab:bodypart}
\end{table}

From~\Cref{tab:bodypart}, we noticed that involving more body parts in the input leads to better performance, despite the intuitive consistency of the head position during movement.

\subsection{Comparison with fusion methods}
\label{fusion}
Current fusion methods improve slightly over silhouette-based methods but are more costly. To analyze their ability to perform under efficiency constraints, we reduce their model size, i.e. Bi-fusion$_{tiny}$. We observe that GaitContour outperforms Bi-fusion$_{tiny}$, demonstrating that GaitContour is a much more \underline{efficient} fusion method.

\begin{table}[h]
\vspace{-3mm}
    \centering
    \caption{Comparison with fusion methods}
    \scalebox{0.8}{
    \begin{tabular}{c c  c c c}
        \hline
        Model & Time(ms) & Gait3D & OUMVLP & GREW\\
        \hline
        Bi-fusion~\cite{peng2023learning} & 79.2 & - & 89.9 & -\\
        Bi-fusion$_{tiny}$ & 55.7 & 13.0 & 7.8 & 5.3\\
        GaitContour & 22.6 & 25.3 & 60.8 & 57.4\\
        \hline
    \end{tabular}
    }
    
    \vspace{-4mm}
\end{table}

To show the superiority of Contour-Pose, we also evaluate its performance when using a large model. SkeletonGait~\cite{fan2023skeletongait} offers a method to transform point representation into an image. We transferred our Contour-Pose to an image format to show that it reserves the most discriminative information.

\begin{table}
    \centering
    \caption{The performance of Contour-Pose transfers to skeleton map on Gait3D. The results are shown using rank retrieval, mean Average Precision(mAP) and mean Inverse
Negative Penalty (mINP)}
    \begin{tabular}{ccccc}
    \toprule
        Method & Rank-1 & Rank-5 & mAP & mINP\\
    \midrule
        SkeletonGait~\cite{fan2023skeletongait} & 38.1 & 56.7 & 28.9 & 16.1 \\
        SkeletonGait++~\cite{fan2023skeletongait} & 77.6 & 89.4 & 70.3 & 42.6\\
        Ours & 77.0 & 89.6 & 69.5 & 47.1\\
    \bottomrule
    \end{tabular}
    \label{tab:skeletongait}
\end{table}

In~\Cref{tab:skeletongait}, Contour-Pose shows comparable performance in Rank retrieval and average precision to SkeletonGait++, while with a 4.5\% improvement in the inverse negative penalty. Our method outperforms SkeletonGait, benefiting from the integration of body shape information and interaction between two modalities. These results collectively highlight Contour-Pose as a representation endowed with rich information and efficiency.

\subsection{Cross-domain evaluation}
\label{cross}
We conduct a cross-domain evaluation to show the generalization ability of GaitContour. We observed that GaitContour is competitive with GPGait, while using significantly fewer parameters. Similarly, when we tailor the GPGait to a similar size as GaitContour, noted GPGait$_{tiny}$, GaitContour would markedly outperform GPGait$_{tiny}$.

\begin{table}[h]
\newcommand{\semitransp}[2][0.35]{\textcolor{fg!#1}{#2}}
\vspace{-3mm}
\caption{Generalization of GaitContour.}
\setlength{\tabcolsep}{2pt}
\scriptsize
    \centering
    \begin{subtable}[t] {.8\linewidth}\centering
    \scalebox{0.9}{\begin{tabular}{c|c|ccc}
    \hline
    Source  & \multirow{2}{*}{Method} & \multicolumn{3}{c}{Testing Datasets}\\
    \cline{3-5}
      Dataset   &  & OUMVLP & GREW & Gait3D\\
    \hline
    \multirow{3}{*}{OUMVLP}  & \textcolor{gray}{\semitransp[20]{GPGait}} & \textcolor{gray}{\semitransp[20]{59.1}} & \textcolor{gray}{\semitransp[20]{11.1}} & \textcolor{gray}{\semitransp[20]{9.0}}\\
         & GPGait$_{tiny}$ & 44.8 & 6.6 & 5.4\\
         & GaitContour & 60.8 & 8.4 & 7.6\\
    \hline
    \multirow{3}{*}{GREW} & \textcolor{gray}{\semitransp[20]{GPGait}} & \textcolor{gray}{\semitransp[20]{4.2}} & \textcolor{gray}{\semitransp[20]{57.0}} & \textcolor{gray}{\semitransp[20]{18.5}}\\
         & GPGait$_{tiny}$ & 2.6 & 28.8 & 14.0\\
         & GaitContour & 4.4 & 57.4 & 16.2\\
    \hline
    \multirow{3}{*}{Gait3D}  & \textcolor{gray}{\semitransp[20]{GPGait}} &
    \textcolor{gray}{\semitransp[20]{2.9}} & \textcolor{gray}{\semitransp[20]{11.0}} & \textcolor{gray}{\semitransp[20]{22.4}}\\
         & GPGait$_{tiny}$ & 1.5 & 6.3 & 17.2 \\
         & GaitContour & 1.8 & 8.9 & 25.3\\
    \hline
    \end{tabular}}
    \end{subtable}
    \begin{subtable}[h] {.8\linewidth}
    \centering
    \scalebox{0.9}{\begin{tabular}{c|c|c}
    \hline
    Model & #Param &  Template \\
    \hline
    GPGait& 7.93 &  $19\times 256$\\
    GPGait$_{tiny}$ & 2.53 & $19\times 128$ \\
    GaitContour & 0.66 \tiny{\textcolor{green}{$\downarrow$ 10X}} &  $1\times 256$  \tiny{\textcolor{green}{$\downarrow$ 19X}} \\

    \hline

    \end{tabular}}
    \end{subtable}
    
    \vspace{-4mm}
\end{table}

\subsection{Contour-Pose++}
\label{++}
A larger model for extracting representation features is one of the reasons why image or fusion-based methods outperform keypoint-based ones. To integrate a large model with Contour-Pose, we follow Skeletongait++~\cite{fan2023skeletongait} by transferring keypoints to image representation. We employ two branches at an early stage to respectively extract silhouette and skeletonmap images, then fuse two sequences features frame-by-frame by an attention mechanism, termed Contour-Pose++. Our Contour-Pose++ outperforms Skeletongait++ by incorporating body shape information in skeletonmap branch, demonstrating the superiority of our proposed Contour-Pose representation.

\begin{figure*}
\centering
\setlength{\tabcolsep}{-1pt}

\begin{tabu}{cccccccc}
\begin{subfigure}{0.08\textwidth}
    \includegraphics[width=\textwidth]{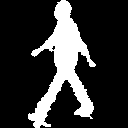}
\end{subfigure} & 
\multirow{2}{*}{
\begin{subfigure}{0.16\textwidth}
    \includegraphics[width=\textwidth, height=\textwidth]{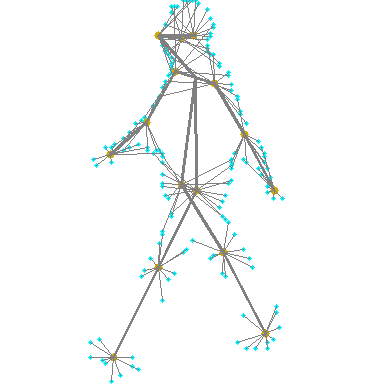}
    \end{subfigure}} &
\begin{subfigure}{0.08\textwidth}
    \includegraphics[width=\textwidth]{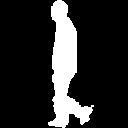}
\end{subfigure} & 
\multirow{2}{*}{
\begin{subfigure}{0.16\textwidth}
    \includegraphics[width=\textwidth, height=\textwidth]{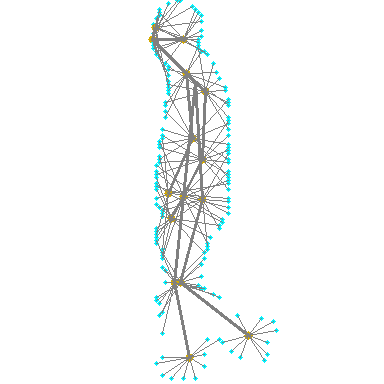}
    \end{subfigure}} &
\begin{subfigure}{0.08\textwidth}
    \includegraphics[width=\textwidth]{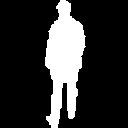}
\end{subfigure} &
\multirow{2}{*}{
\begin{subfigure}{0.16\textwidth}
    \includegraphics[width=\textwidth, height=\textwidth]{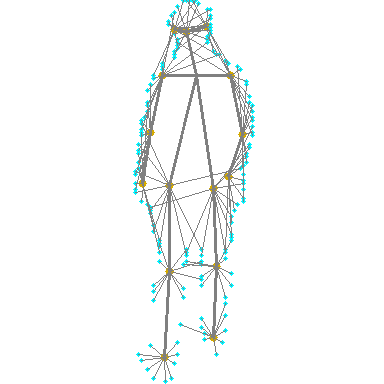}
    \end{subfigure}}&
\begin{subfigure}{0.08\textwidth}
    \includegraphics[width=\textwidth]{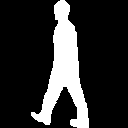}
\end{subfigure} & 
\multirow{2}{*}{
\begin{subfigure}{0.16\textwidth}
    \includegraphics[width=\textwidth, height=\textwidth]{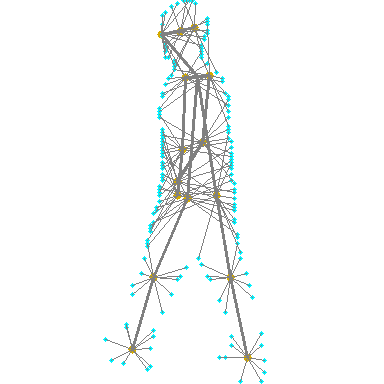}
    \end{subfigure}}     
    \\ 
\begin{subfigure}{0.08\textwidth}
    \includegraphics[width=\textwidth, height=\textwidth]{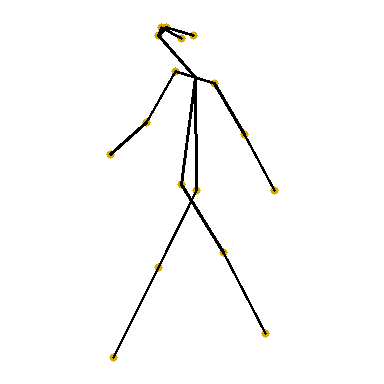}
\end{subfigure}  &  &
\begin{subfigure}{0.08\textwidth}
    \includegraphics[width=\textwidth, height=\textwidth]{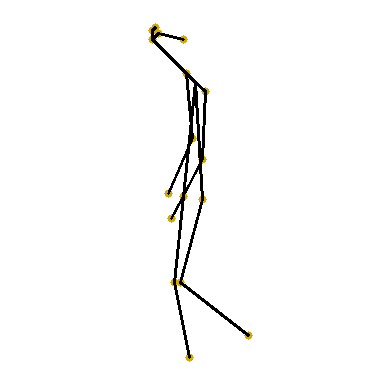}
\end{subfigure} &  &
\begin{subfigure}{0.08\textwidth}
    \includegraphics[width=\textwidth, height=\textwidth]{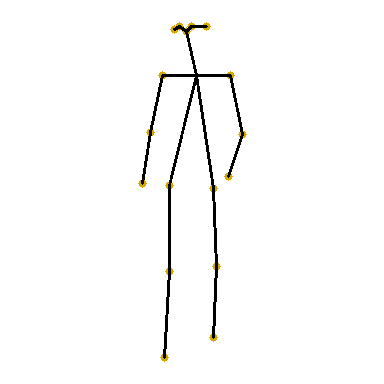}
\end{subfigure}  & & 
\begin{subfigure}{0.08\textwidth}
    \includegraphics[width=\textwidth, height=\textwidth]{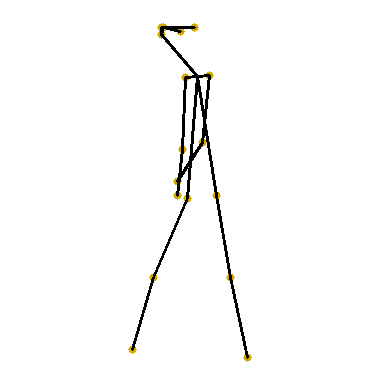}
\end{subfigure}  &\\
\end{tabu}
\caption*{OUMVLP}

\begin{tabu}{cccccccc}
\begin{subfigure}{0.08\textwidth}
    \includegraphics[width=\textwidth]{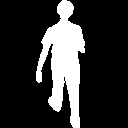}
\end{subfigure} & 
\multirow{2}{*}{
\begin{subfigure}{0.16\textwidth}
    \includegraphics[width=\textwidth, height=\textwidth]{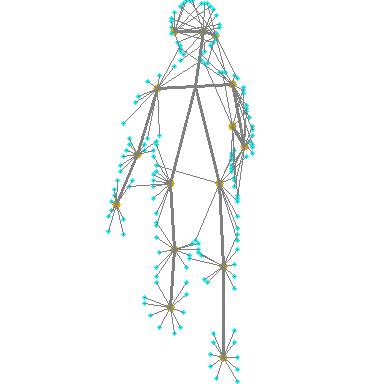}
    \end{subfigure}} &
\begin{subfigure}{0.08\textwidth}
    \includegraphics[width=\textwidth]{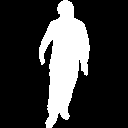}
\end{subfigure} & 
\multirow{2}{*}{
\begin{subfigure}{0.16\textwidth}
    \includegraphics[width=\textwidth, height=\textwidth]{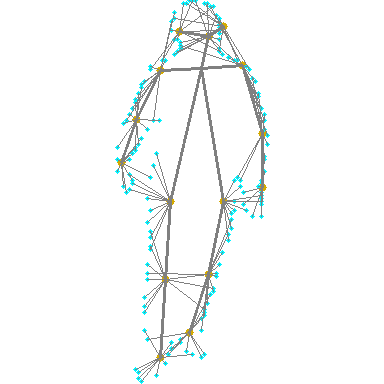}
    \end{subfigure}} &
\begin{subfigure}{0.08\textwidth}
    \includegraphics[width=\textwidth]{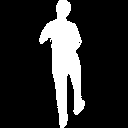}
\end{subfigure} &
\multirow{2}{*}{
\begin{subfigure}{0.16\textwidth}
    \includegraphics[width=\textwidth, height=\textwidth]{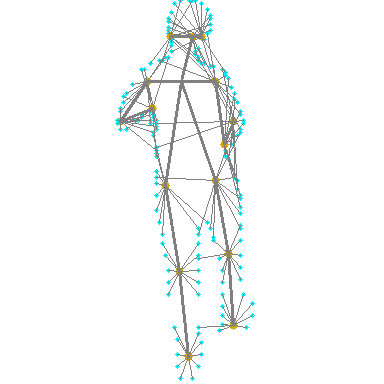}
    \end{subfigure}}&
\begin{subfigure}{0.08\textwidth}
    \includegraphics[width=\textwidth]{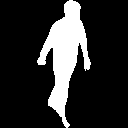}
\end{subfigure} & 
\multirow{2}{*}{
\begin{subfigure}{0.16\textwidth}
    \includegraphics[width=\textwidth, height=\textwidth]{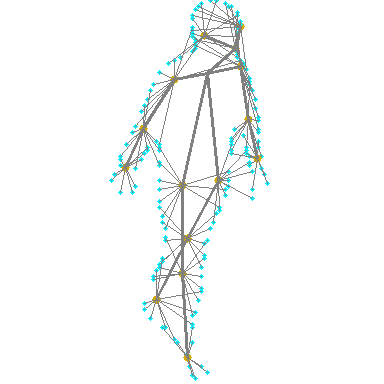}
    \end{subfigure}}     
    \\ 
\begin{subfigure}{0.08\textwidth}
    \includegraphics[width=\textwidth, height=\textwidth]{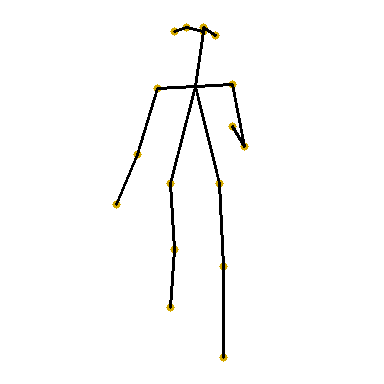}
\end{subfigure}  &  &
\begin{subfigure}{0.08\textwidth}
    \includegraphics[width=\textwidth, height=\textwidth]{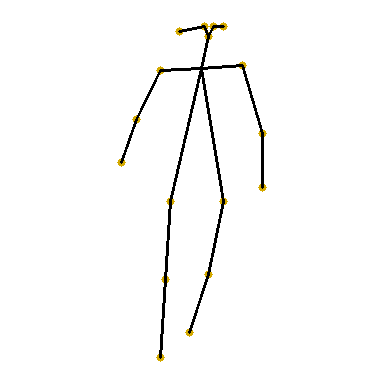}
\end{subfigure} &  &
\begin{subfigure}{0.08\textwidth}
    \includegraphics[width=\textwidth, height=\textwidth]{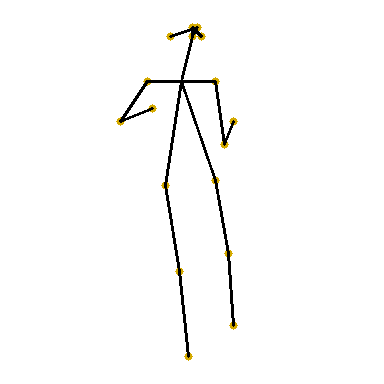}
\end{subfigure}  & & 
\begin{subfigure}{0.08\textwidth}
    \includegraphics[width=\textwidth, height=\textwidth]{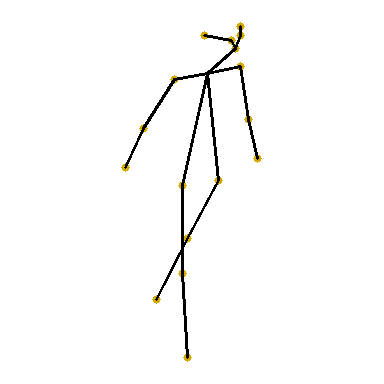}
\end{subfigure}  &\\
\end{tabu}
\caption*{Gait3D}

\begin{tabu}{cccccccc}
\begin{subfigure}{0.08\textwidth}
    \includegraphics[width=\textwidth]{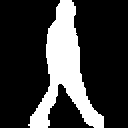}
\end{subfigure} & 
\multirow{2}{*}{
\begin{subfigure}{0.16\textwidth}
    \includegraphics[width=\textwidth, height=\textwidth]{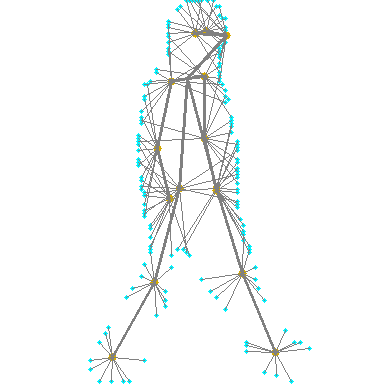}
    \end{subfigure}} &
\begin{subfigure}{0.08\textwidth}
    \includegraphics[width=\textwidth]{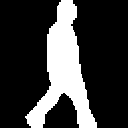}
\end{subfigure} & 
\multirow{2}{*}{
\begin{subfigure}{0.16\textwidth}
    \includegraphics[width=\textwidth, height=\textwidth]{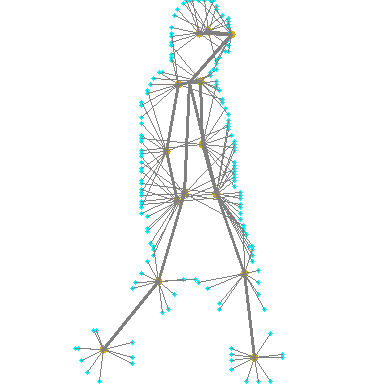}
    \end{subfigure}} &
\begin{subfigure}{0.08\textwidth}
    \includegraphics[width=\textwidth]{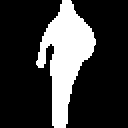}
\end{subfigure} &
\multirow{2}{*}{
\begin{subfigure}{0.16\textwidth}
    \includegraphics[width=\textwidth, height=\textwidth]{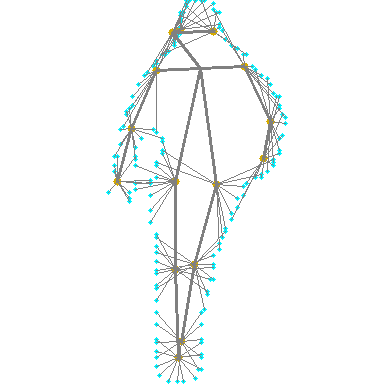}
    \end{subfigure}}&
\begin{subfigure}{0.08\textwidth}
    \includegraphics[width=\textwidth]{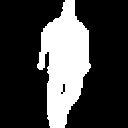}
\end{subfigure} & 
\multirow{2}{*}{
\begin{subfigure}{0.16\textwidth}
    \includegraphics[width=\textwidth, height=\textwidth]{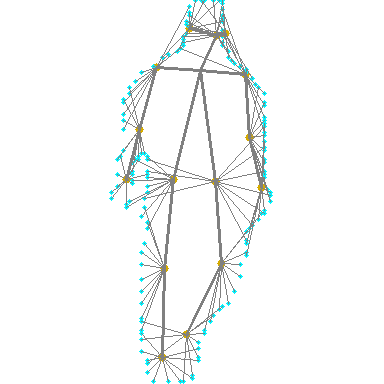}
    \end{subfigure}}     
    \\ 
\begin{subfigure}{0.08\textwidth}
    \includegraphics[width=\textwidth, height=\textwidth]{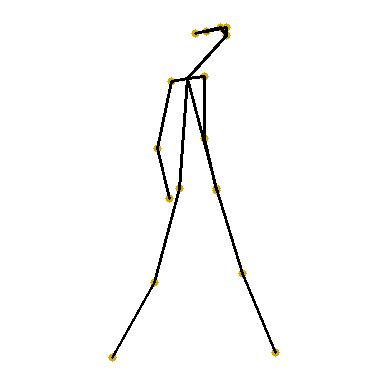}
\end{subfigure}  &  &
\begin{subfigure}{0.08\textwidth}
    \includegraphics[width=\textwidth, height=\textwidth]{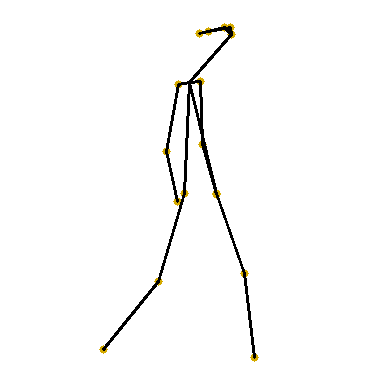}
\end{subfigure} &  &
\begin{subfigure}{0.08\textwidth}
    \includegraphics[width=\textwidth, height=\textwidth]{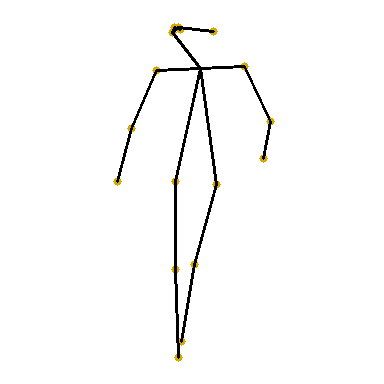}
\end{subfigure}  & & 
\begin{subfigure}{0.08\textwidth}
    \includegraphics[width=\textwidth, height=\textwidth]{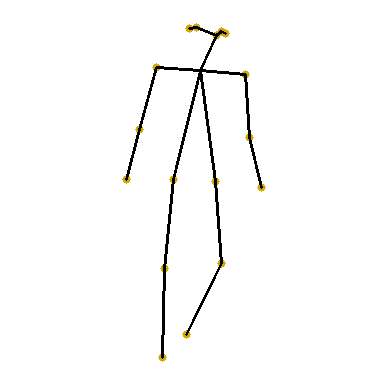}
\end{subfigure}  \\
\end{tabu}
\caption*{GREW}

\begin{tabu}{cccccccc}
\begin{subfigure}{0.08\textwidth}
    \includegraphics[width=\textwidth]{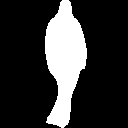}
\end{subfigure} & 
\multirow{2}{*}{
\begin{subfigure}{0.16\textwidth}
    \includegraphics[width=\textwidth, height=\textwidth]{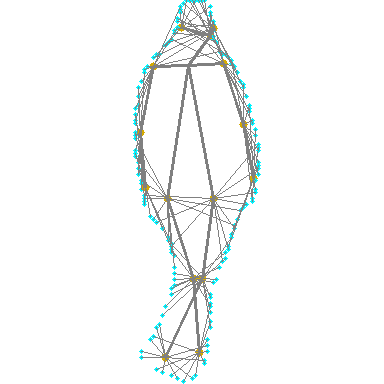}
    \end{subfigure}} &
\begin{subfigure}{0.08\textwidth}
    \includegraphics[width=\textwidth]{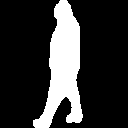}
\end{subfigure} & 
\multirow{2}{*}{
\begin{subfigure}{0.16\textwidth}
    \includegraphics[width=\textwidth, height=\textwidth]{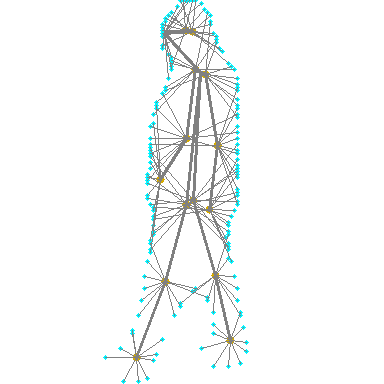}
    \end{subfigure}} &
\begin{subfigure}{0.08\textwidth}
    \includegraphics[width=\textwidth]{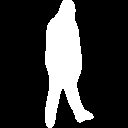}
\end{subfigure} &
\multirow{2}{*}{
\begin{subfigure}{0.16\textwidth}
    \includegraphics[width=\textwidth, height=\textwidth]{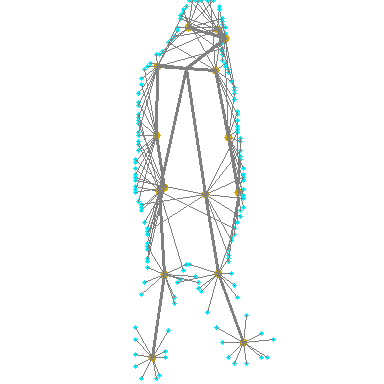}
    \end{subfigure}}&
\begin{subfigure}{0.08\textwidth}
    \includegraphics[width=\textwidth]{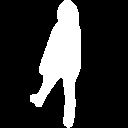}
\end{subfigure} & 
\multirow{2}{*}{
\begin{subfigure}{0.16\textwidth}
    \includegraphics[width=\textwidth, height=\textwidth]{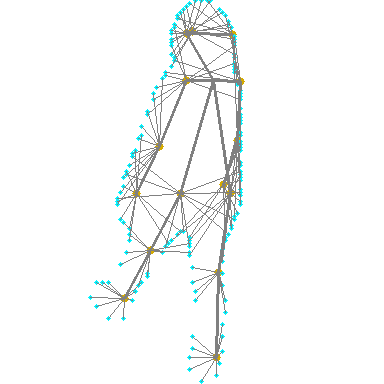}
    \end{subfigure}}     
    \\ 
\begin{subfigure}{0.08\textwidth}
    \includegraphics[width=\textwidth, height=\textwidth]{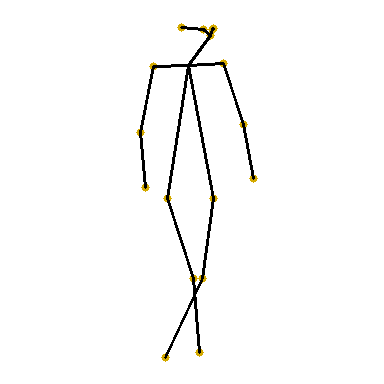}
\end{subfigure}  &  &
\begin{subfigure}{0.08\textwidth}
    \includegraphics[width=\textwidth, height=\textwidth]{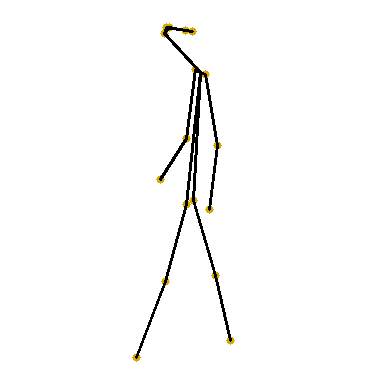}
\end{subfigure} &  &
\begin{subfigure}{0.08\textwidth}
    \includegraphics[width=\textwidth, height=\textwidth]{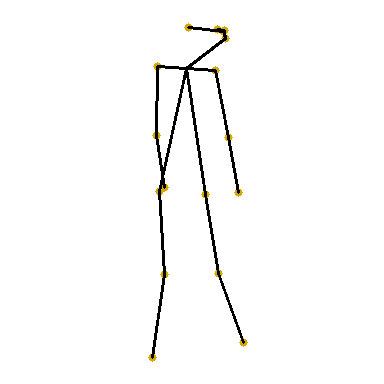}
\end{subfigure}  & & 
\begin{subfigure}{0.08\textwidth}
    \includegraphics[width=\textwidth, height=\textwidth]{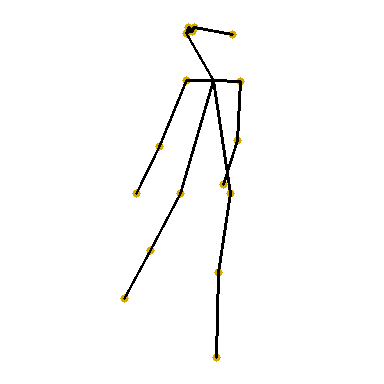}
\end{subfigure}  &\\
\end{tabu}
\caption*{BRIAR}

\caption{Contour-Pose visualization on multiple gait recognition datasets.}
\label{fig:visualization}
\end{figure*}

\newpage
{
    \bibliographystyle{splncs04}
    \bibliography{main}
}